\documentclass{article}

\usepackage[preprint, nonatbib]{neurips_2026}

\usepackage[utf8]{inputenc} 
\usepackage[T1]{fontenc}    
\usepackage{hyperref}       
\usepackage{url}            
\usepackage{booktabs}       
\usepackage{amsfonts}       
\usepackage{nicefrac}       
\usepackage{microtype}      
\usepackage{xcolor}         
\usepackage{amsmath}
\usepackage{booktabs}
\usepackage{tabularx}
\usepackage{array}
\usepackage{graphicx}
\usepackage{rotating}
\usepackage{placeins}
\usepackage[style=numeric, backend=biber]{biblatex} 
\title{ConnectomeBench2: A Unified Benchmark for Automated Connectomic Proofreading}

\author{%
  Jeff Brown \thanks{indicates equal contribution} \\
  MIT\\
  \texttt{jffbrwn@mit.edu} 
  \And
  Tim Farkas$^*$ \\
  MIT\\
  \\
  \And
  Gleb Razgar \\
  The Open University\\
  \And
  Edward S. Boyden \\
  Mindspan Institute\\
  Yang Tan Collective\\
  McGovern Institute\\
  MIT Departments of Brain and Cognitive Sciences and Biological Engineering\\
  \texttt{edboyden@mit.edu} 
}

\begin{document}

\maketitle

\begin{abstract}
Proofreading—correcting segmentation errors in 3D brain reconstructions—is the rate-limiting step in synapse-resolution connectomics. We release ConnectomeBench2, a unified multi-species dataset of over 716,485 expert-labeled proofreading decisions with >4,500,000 associated images spanning four major open connectomes (mouse, human, zebrafish, fly), spanning both split and merge error correction. Trained on this dataset, a single Vision Transformer with shared encoders for mesh geometry and electron microscopy reaches human-level accuracy across species for split error correction and merge error identification, with performance scaling with data size and modality. Beyond accuracy, we show that the model is well-calibrated within distribution, that measures of distribution distance predict where calibration and accuracy will degrade on unseen data, and that connectomics-specific pretraining and active learning-based sample selection show potential to substantially reduce the labeling effort needed to extend to new species and brain regions. The benchmark provides the infrastructure to train and evaluate increasingly capable vision models for connectomic proofreading.

\paragraph{Data and code availability.} The ConnectomeBench2 dataset is released on Hugging Face at \url{https://huggingface.co/datasets/jeffbbrown2/ConnectomeBench2}. The accompanying codebase is available on GitHub at
\url{https://github.com/timfarkas/ConnectomeBench2}.
\end{abstract}

\section{Introduction}
Connectomics aims to map the structure of nervous systems at the level of individual neurons and their connections~\cite{sporns2005human, microns2025functional, dorkenwald2024flywire}. To produce these maps, brain tissue is imaged at high resolution using electron or expansion microscopy~\cite{tavakoliLightmicroscopybasedConnectomicReconstruction2025}. 3D segmentation algorithms partition the imaging volumes into biological structures such as neurons, glia, and blood vessels~\cite{januszewski2018ffn, sheridan2023lsd}. State-of-the-art segmentation algorithms have improved substantially over the years, but they still introduce errors that must be manually corrected afterward, a process known as ``proofreading.'' There are two types of errors corrected during proofreading: split errors, when a single neuron's segment is falsely split into multiple pieces; and merge errors, where the segments of multiple neurons are fused together. The goal of proofreading is to remove these errors, trace out complete neurons, and annotate connections between neurons (e.g., synapses). Given that it is highly time-intensive~\cite{dorkenwald2022flywire, dorkenwald2024flywire}, proofreading remains the primary bottleneck in generating synapse-resolution connectivity maps of whole brains.

Many automated proofreading approaches have been proposed to accelerate this process ~\cite{celii2025neurd, januszewski2025pathfinder, haehn2017guided, berman2022bridging, troidl2024pointaffinity, rolnick2017morphological, li2020subcompartment, matejek2019biologically, schubert2019cmn, schmidt2024roboem, nguyen2021rlcorrector, joyce2023mesh, meirovitch2016multipass, sicat2013graph, plaza2016focused, svara2022zebrafish}. We seek to extend on this prior work by leveraging key lessons of modern machine learning (ML): Namely that, (i) simple methods that scale often yield strong performance; (ii) multi-task and joint training across related distributions often outperform siloed per-task or per-distribution models; and (iii) pretrained representations transfer efficiently to new data. Furthermore, on the evaluation side, deployment-ready systems are evaluated not just on accuracy but on calibration of their uncertainty, on detection of out-of-distribution (OOD) inputs, and on sample efficiency when adapting to new distributions. 

Our work thus makes three contributions:
\begin{enumerate}
    \item \textbf{A labeled multi-task, multi-species dataset of >4,500,000 proofreading samples.} We establish a unified visual formulation of proofreading drawing on expert proofreader labels extracted from the four major open-source connectomics projects in mouse, human, zebrafish, and fly.
    \item \textbf{Human-level accuracy in split error correction and false merge identification} in unified vision models trained and evaluated on the dataset. Performance scales with data size and diversity.
    \item \textbf{Strong calibration and comprehensive evaluation of generalization.} We show that models are well-calibrated within their training distribution, and that measures of distribution shift can be used to predict reduction of performance and calibration. We demonstrate that strategic sample selection improves generalization to unseen data.
\end{enumerate}

Together, these results establish that bringing modern training and evaluation practices to proofreading yields concrete deployment-relevant gains, and that the benchmark provides the infrastructure for the field to continue extending them.

\begin{figure}[h]
     \centering
    \includegraphics[width=\linewidth]{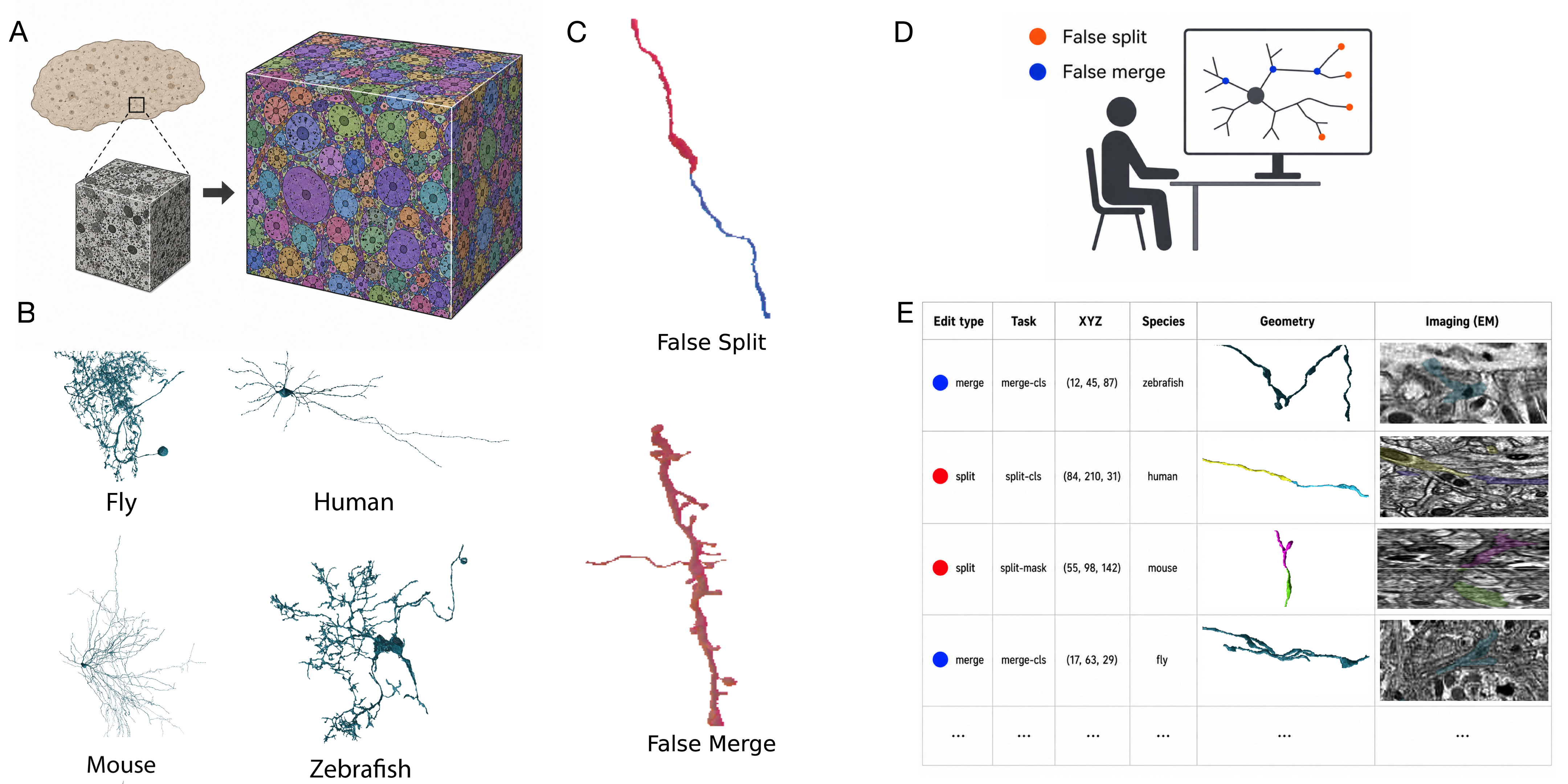}
    \caption{Connectomic proofreading, and leveraging expert proofreading labels for machine learning.
    A: Scanning and segmentation of brain tissue (Cartoon). B: Exemplary traced neuron meshes of each of the four connectomes included. C: Examples of Split Error and Merge Error. D: Manual Proofreading Community Effort. E: Structured extraction of manual edits for connectomic proofreading model training and evals.}
    \label{fig:mylabel}
\end{figure}
\section{Related Work}


\paragraph{Automated Proofreading methods.} A range of automated proofreading systems have been developed, varying along several axes. Most use machine learning, with 3D CNNs being the most common architectural choice~\cite{zung2017error, rolnick2017morphological, li2020subcompartment, matejek2019biologically, huang2025autoproof}, alongside 2D CNN approaches operating on projections or context regions~\cite{haehn2017guided, schubert2019cmn}, or point-cloud methods~\cite{berman2022bridging, troidl2024pointaffinity}. Other methods avoid machine learning entirely, applying hand-crafted rules over geometric or graph representations~\cite{celii2025neurd, joyce2023mesh, meirovitch2016multipass, sicat2013graph, plaza2016focused}. These methods also differ in which errors they target. Some address only merge errors~\cite{celii2025neurd, rolnick2017morphological, li2020subcompartment, schubert2019cmn, meirovitch2016multipass}, others only splits~\cite{schmidt2024roboem, matejek2019biologically, berman2022bridging, joyce2023mesh, chen2024multimodal, huang2025autoproof}, and a smaller set handle both~\cite{zung2017error, haehn2017guided, nguyen2021rlcorrector, troidl2024pointaffinity, januszewski2025pathfinder, sicat2013graph, plaza2016focused}, though typically as separate sub-models. Inputs vary similarly: some methods operate purely on segmentation derivatives such as meshes or skeletons~\cite{celii2025neurd, januszewski2025pathfinder, joyce2023mesh, berman2022bridging, troidl2024pointaffinity}, others on raw imaging volumes~\cite{schmidt2024roboem}, and a third class fuses both~\cite{zung2017error, haehn2017guided, li2020subcompartment, matejek2019biologically, chen2024multimodal, huang2025autoproof}. 

\paragraph{Species coverage.} Most prior systems are evaluated on a narrow species range. Several focus on \emph{Drosophila}~\cite{huang2025autoproof, chen2024multimodal, troidl2024pointaffinity, plaza2016focused}, others on mouse cortex~\cite{januszewski2025pathfinder}, and a few extend to multiple species: NEURD~\cite{celii2025neurd} and RoboEM~\cite{schmidt2024roboem} cover mouse and human cortex, and Berman et al.~\cite{berman2022bridging} demonstrate the missing-section problem across fly, mouse, and human while training only on fly data. Svara et al.~\cite{svara2022zebrafish} reconstruct the larval zebrafish brain. Cross-species transfer has been studied for segmentation~\cite{sawmya2025neuroaddaactivediscriminativedomain}, but to our knowledge no prior proofreading system has been trained or evaluated across the full breadth of species available in modern connectomics.

\paragraph{Closest comparisons.} Two recent efforts most closely relate to ours. Autoproof~\cite{huang2025autoproof} scales training data for proofreading models, demonstrating that careful data curation enables strong performance, but is restricted to \emph{Drosophila} and to merge-error decisions on focused fragment pairs. ConnectomeBench~\cite{brown2025connectomebenchllmsproofreadconnectome} introduces an evaluation benchmark for proofreading systems across two species, sharing our motivation that the field needs standardized evaluation. However, its size (hundreds of samples) is too small to support model training, and its merge-error evaluation compares predictions against fully proofread neuron meshes rather than local corrections, which can produce ambiguous ground truth. Our work extends both: a dataset large enough to both train and evaluate on, that spans four species, supports both error types under consistent task definitions, and is curated to ensure evaluation reflects the local correction being made.

\section{Data}

\subsection{Data Access and Generation}
One of the core objectives of this work was the comprehensive extraction and organization of data from major public connectomics proofreading efforts (FlyWire \cite{dorkenwald2024flywire,dorkenwald2022flywire}, MICrONS \cite{microns2025functional}, Fish1 \cite{Petkova2025.06.10.658982}, H01 \cite{shapsoncoe2024human}). The MICrONS dataset (microns-explorer.org) and the H01 dataset (h01-release.storage.googleapis.com) are released under the Creative Commons Attribution 4.0 International license (CC BY 4.0). The FlyWire dataset (flywire.ai) and the Fish1 dataset (fish1-release.storage.googleapis.com) are released under the Creative Commons Attribution-NonCommercial 4.0 International license (CC BY-NC 4.0). We acknowledge their efforts and release our dataset with a license that respects the licenses of these four datasets.

\subsubsection{Raw Connectomics Source Data}
Connectomics data was downloaded from host servers through CloudVolume \cite{10.3389/fncir.2022.977700}. 
We used CAVEClient \cite{dorkenwaldCAVEConnectomeAnnotation2025} to fetch the proofreading segmentation graph.  
This graph consists of two abstractions: 1. the supervoxel, a set of imaging voxels belonging to the same small local segment, and 2. the root, a set of supervoxels grouped at a specific timestamp that represents a unique, putative neuron segment.
Roots are nodes in a directed proofreading edit lineage graph: A \textit{merge edit} (that corrects a split error or false split) creates a new root subsuming two or more previous ones. A \textit{split edit} (that corrects a merge error or false merge) creates two or more new roots out of one previous root.
For mouse and fly, we collect proofreading edits from proofread roots fetched from these species'
respective proofreading status tables (\texttt{proofreading\_status\_and\_strategy} for mouse, \texttt{proofread\_neurons} for fly). Human and zebrafish do not maintain such tables, so for these species we sample edits from all roots that received proofreading edits.

\subsubsection{Generating Positive Examples}
\paragraph{Operation Table} Starting from each species' final proofread roots, we traverse the proofreading graph up to the first ancestor roots, and add any unique edits to a structured table of proofreading operations (see Appendix \ref{sec:appendix-data-sources} for details). For each edit, we store its operation ID (primary key in CaveClient), edit type (merge or split), 3D coordinates (average of the sink and source points the human proofreader placed to specify the operation), and the involved root ids (before and after). This single 3D coordinate is also what determines the edit's assignment to a spatial cube during dataset splitting (see below). To get high quality proofreading actions, only operations that were not later reversed are included. Additionally, we excluded segments too small for meaningful proofreading decisions via morphology; if the proofreading actions corrected a split error (combined two segments), we kept both segments when they met species-specific size thresholds hand-tuned to exclude small dust. 

\subsubsection{Generating Negative Examples}
We use three strategies to generate labeled samples of plausible-but-false samples for classification tasks (see Section \ref{sec:methods-tasks} below)

\textbf{Negative Split Error Corrections.} We use two kinds of candidates that are plausible but incorrect: 
\begin{itemize}
    \item \textbf{Synapses between two proofread neurons.} These are controls where the model is asked whether the two synapsing segments should be merged; the correct answer is no, since by construction they belong to different neurons. Only fly and mouse synapses were available from CAVEClient (see Appendix \ref{sec:appendix-synapses}) \cite{buhmann2021automatic, heinrich2018synaptic, turner2020synaptic}. From here on, we call these synapse controls. Furthermore, these are used for an auxiliary synapse classification task.
    \item \textbf{Adjacent segments} To generate alternative candidates, we sample adjacent segments within a 200nm bounding box of the 3D coordinate where the edit occurred, excluding the true merge candidate and dust. These are included because split error correction requires distinguishing the correct continuation of a false split from adjacent bad continuations.
\end{itemize} 
\paragraph{Negative Merge Errors}
A plausible but bad candidate for a false merge correction is a site that is known to not contain false merges (i.e., that is error-free), yet cannot be trivially distinguished from false merge sites. To achieve this, we sample junctions of proofread roots' skeletons (i.e., nodes of degree $>$2), as false merges commonly manifest as spurious junctions (for more details, see Appendix \ref{sec:appendix-junctions}). From here on, we call these junction controls.

\subsubsection{Data Splitting}
The proofread volume was divided into ((80$\mu\mathrm{m})^3$) cubes (mouse: 1,960, fly: 264, human: 5,616, zebrafish: 784), randomly assigned to one of the train, validation, and test splits (75/12.5/12.5). Edit operations and synapse and junction controls are assigned to the split of the cube that their associated 3D coordinates fall in.
In other words, splitting was done spatially rather than on a neuron identity or edit level. This is because intermediate roots often cannot be unambiguously assigned to any one final root and to rule out leaks from semantically correlated edits in close proximity (e.g.\ a split followed by a merge). Considering near-duplicate edit-site leakage across cube boundaries, the fraction of val/test proofreading edits within 500nm of a training edit across the cube split boundary was <0.5\% across species, which we deemed negligible.

\subsubsection{Data Rendering}
For each sample, we render two modalities: mesh geometry and imaging. (See Appendix \ref{sec:appendix-data-rendering} for details.)

\textbf{Mesh geometry} views are 2D orthographic renderings of mesh segments in three spatial views (top, side, and front), where each view consists of 7 channels encoding geometric information: three normals, depth, two segment mask channels, and silhouette (see Appendix Figure ~\ref{fig:geo-em-data-example-rendering}). Assignment of segments to mask channels A and B follows the return order from the segmentation graph.

\textbf{Imaging} views are four images, each showing a different 2D slice: cardinal slices XY, XZ, YZ, and one oblique slice picked to maximize visible segmentation area (log sum of segment areas). Each is rendered in 3 channels, with the first being the microscopy imaging data (electron microscopy, EM), and the second and third being the segmentation mask of the two segments (see Appendix Figure ~\ref{fig:geo-em-data-example-rendering}). The second mask is all-zeros in single-segment tasks.

\subsubsection{Dataset Distribution}
While we aimed to keep labels within tasks relatively balanced (roughly 1:1), our dataset is imbalanced in both species and tasks (cf. Figure \ref{fig:dataset_sizes}) for several reasons: 1) number of available qualifying proofreading edits (across species and tasks), 2) availability of controls/synapses (across species), 3) rendering/download time, especially for datasets with larger roots (across species), 4) time/network throughput constraints.

\subsection{Model Training}
\subsubsection{Task Framing}
\label{sec:methods-tasks}
We frame connectomic proofreading as essentially two tasks: 1) identifying whether two neuron segments are the same neuron and must be merged; and 2) whether a given segment contains a false merge that must be split. Correcting a false split means merging in the correct merge partner, a binary classification task. Correcting a false merge, on the other hand, requires identifying an exact split boundary.
We thus frame this as three tasks for our model:
\begin{enumerate}
\item Split Error Correction
\item Merge Error Classification
\item Mask Segmentation for Merge Error Correction
\end{enumerate}
The compiled dataset here presented allows supervised training on each of these tasks with a unified input format (7-channel geometry format and 3-channel EM format).

\paragraph{Model.} The unified model is a single Vision Transformer
(ViT-B/16 or ViT-L/16, $224\!\times\!224$ inputs) with \emph{two} patch
embeddings sharing one encoder: a 7-channel module for geometry views and a
3-channel module for imaging views, both initialized from ImageNet weights
(RGB filters copied into the three normal channels for geometry; remaining
geometry channels Kaiming-initialized). Three classification heads on the
CLS token handle split error correction, merge error identification, and an
auxiliary synapse decision; a CNN decoder on the patch tokens upsamples to
$224\!\times\!224$ and outputs a 2-channel mask for merge error correction.
At inference time, view-level probabilities are averaged within each
operation to produce the per-sample (per-op) decision and confidence.

\paragraph{Joint multi-task loss.} 
\label{sec:task-loss} Each batch mixes views across species,
modalities, and tasks; every head produces an output for every sample, and
losses are masked to the relevant head per sample. The classification heads
use softmax cross-entropy with label smoothing $0.1$. The mask loss is
\emph{permutation-invariant} over the two output channels, since the A/B
labeling of the two pieces produced by a split is arbitrary:
\[
\mathcal{L}_{\text{mask}}
=
\min_{\pi \in \{\text{id}, \text{swap}\}}\!
\Big[\, w_{\text{CE}}\,\mathrm{CE}^{\pi}
+ (1{-}w_{\text{CE}})(1{-}\overline{\mathrm{Dice}}^{\pi})\,\Big],
\quad w_{\text{CE}}{=}0.5,
\]
computed on foreground pixels only and gated to rows with valid dual-mesh
ground-truth. $\text{CE}$ is pixel-level cross entropy loss; $\overline{\mathrm{Dice}}$ is mean Dice Loss \cite{milletari2016vnetfullyconvolutionalneural}. The total loss is
$\mathcal{L}_{\text{ep}} + \mathcal{L}_{\text{jn}} + \mathcal{L}_{\text{mask}}
+ 0.25\,\mathcal{L}_{\text{syn}}$, with the synapse classification loss term down-weighted as
auxiliary.

\paragraph{Training.} We train with AdamW + cosine schedule on 
H100 GPUs (1--3 GPUs per job, batch size 256--768), capped at 24 GPU-hours
per run. The peak ViT-B run completes in $\sim$15 GPU-hours; ViT-L on
3$\times$H100 in $\sim$20. Inference runs on a single A100 in bfloat16.
Full optimizer, augmentation, blending, and seed details are in
Appendix~\ref{sec:appendix-reproducibility}.

\section{Accuracy: How well does the system correct errors?}

\begin{table*}[t]
\centering
\caption{Peak performance and ablations on the held-out test set. Per-sample (one prediction per operation; views aggregated). bAcc is reported at threshold 0.5 alongside mIoU (mask GT is rendered on geometry views only, so the EM-only row's mIoU is `—'). Brackets are 95\% cluster-bootstrap confidence intervals.}
\label{tab:peak_performance}
\begin{tabular}{lccc}
\toprule
Model & Split Error bAcc (\%) & Merge Error bAcc (\%) & mIoU \\
      & [95\% CI]              & [95\% CI]              & [95\% CI] \\
\midrule
\multicolumn{4}{l}{\textit{Baseline}} \\
Human Experts & 93.0 [87.2, 97.7]  & 84.1 [75.8, 91.7]  & — \\

\midrule
\multicolumn{4}{l}{\textit{Architecture}} \\
ViT-L & 97.2 [97.1, 97.4] & 92.8 [92.5, 93.0] & 0.678 [0.676, 0.681] \\
ViT-B  & 97.0 [96.9, 97.2] & 93.0 [92.7, 93.2] & 0.696 [0.693, 0.698] \\
\midrule
\multicolumn{4}{l}{\textit{Modality \& channel ablations on ViT-B (full)}} \\
\quad ViT-B, geom only & 96.2 [96.0, 96.4] & 90.6 [90.4, 90.9] & 0.696 [0.693, 0.698] \\
\quad $-$shape & 96.2 [96.0, 96.4] & 89.7 [89.4, 90.0] & 0.677 [0.675, 0.680] \\
\quad $-$normals & 96.0 [95.9, 96.2] & 88.2 [87.9, 88.5] & 0.680 [0.677, 0.683] \\
\quad $-$masks  & 77.7 [77.4, 78.1] & 90.6 [90.4, 90.9] & 0.696 [0.693, 0.698] \\
\quad ViT-B, EM only & 94.2 [94.0, 94.4] & 91.1 [90.9, 91.4] & — \\
\bottomrule
\end{tabular}
\end{table*}

The first property relevant to any proofreading system is accuracy: how often does the system correct errors versus introduce them? In Table~\ref{tab:peak_performance}, we see that the ViT-B reaches human-level performance on split error correction and merge error identification. On detecting synapses, performance is strong on both mouse and fly synapses, with per-image balanced accuracy of 98.2\% [98.1, 98.4] on mouse (TPR 98.5\%, TNR 98.0\%) and 96.1\% [95.8, 96.4] on fly (TPR 94.5\%, TNR 97.7\%) (see Table 13). For the merge error correction task, we measure performance as mask Intersection over Union (mIoU, demonstrated in \ref{fig:interpretability}, right). The relationship between mIoU and split quality is shown in Appendix Figure ~\ref{fig:sv-iou-vs-mask-iou}.


\begin{figure}[h]
    \centering
    \includegraphics[width=\textwidth]{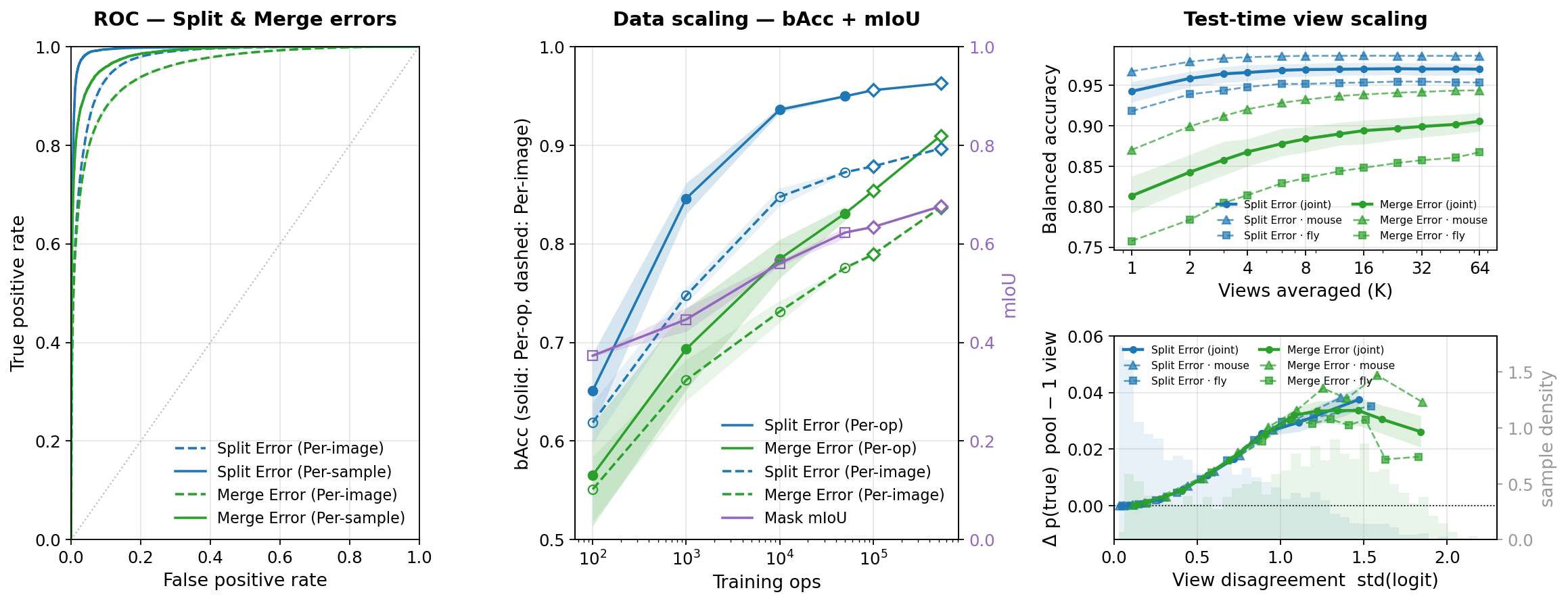}
    \caption{\textbf{(a)} ROC curve for the joint Split-Error / Merge-Error classifier on the held-out test set. Per-sample rows aggregate views to one prediction per operation; Per-image rows score each view independently. Brackets are 95\% cluster-bootstrap confidence intervals. \textbf{(b)} Balanced accuracy (left axis) and mIoU (right axis) versus number of training operations on a log $x$ axis. bAcc is shown at two granularities: Per-op (solid; mean view probability per operation, thresholded at 0.5) and Per-image (dashed). Each curve is the unweighted mean across the four species. Bands are $\pm$1 SD across training seeds. Separately seeded runs per scale: 5 at 100, 3 at 1,000, 3 at 10,000, 2 at 50,000, and 1 at 100,000 and all samples. Single-seed points are drawn as diamonds without bands. 
    \textbf{(c)} Test-time view scaling. \emph{Top:} balanced accuracy of the geometric (mean-logit) pool versus the number of rendered geometry views averaged, $K$ (log axis); the metric is computed per single-$K$ realization then averaged over random $K$-subsets, with op-clustered bootstrap 95\% CI bands. \emph{Bottom:} per-operation pooling benefit $\Delta p(\text{true})$ (64-view pool minus a single view) versus per-op view disagreement (std of the per-view logits), with the sample-density histogram behind.
}
    \label{fig:scaling_and_species}
\end{figure}

Having established overall performance, we next ask which aspects of the system drive it. We run a series of ablations on the ViT family, with results shown in Table~\ref{tab:peak_performance}. Because our rendering format separates mesh normals, depth, segmentation masks, and imaging into distinct channels, we can ablate each input modality independently at inference time. While most ablations do not dramatically effect performance (likely due to channel dropout during training), we find that removing masks from split error correction tasks substantially impacts performance. 

Next, we examine the effect of training data scale, both in dataset size and diversity of tasks, species, and number of views. Balanced accuracy improves consistently with training data across all three tasks (Figure~\ref{fig:scaling_and_species}, Middle). 
More notably, we find performance to increase with the number of views sampled around a proofreading site and aggregated into a prediction. Beyond seeing strong improvements in TPR/TNR when going from single-image predictions to aggregated predictions of all seven views per sample present in our dataset (3 geometry views, 4 imaging views, Figure~\ref{fig:scaling_and_species}, Left), we saw performance scaling further with additionally sampled views (different additional rotations and extents): Aggregating predictions across up to 64 geometry views recovered up to ~3-4\% percentage points of accuracy. The highest increase was observed in edits with higher disagreement between views. This view scaling was observed across the two binary tasks and species tested (mouse and fly). In proofreading deployment, this dynamic may allow for adaptive sampling to trade off more compute for better performance in complex cases with higher view disagreement, particularly for the merge error identification task, where the fraction of high-disagreement samples is higher. (Figure~\ref{fig:scaling_and_species}, Right)

\begin{figure}[h]
    \centering
    \includegraphics[width=\textwidth]{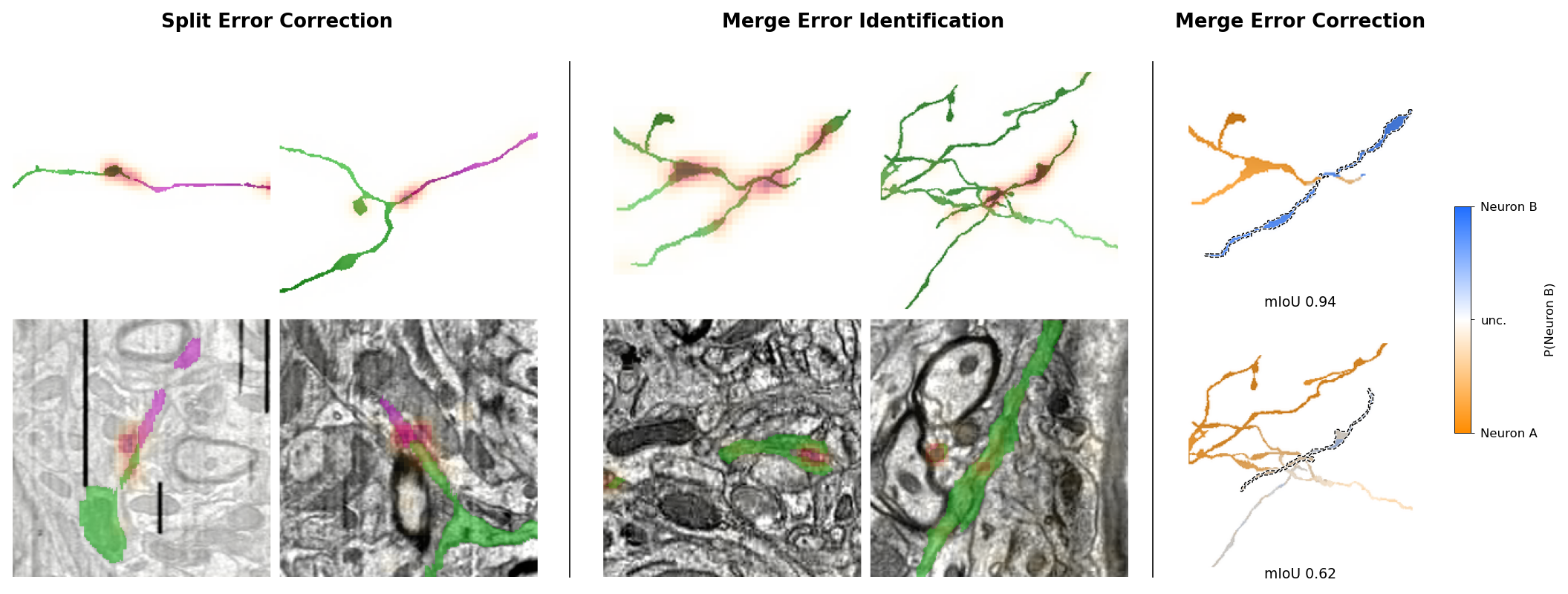}
    \caption{Qualitative examples on held-out examples from the test set.
\textbf{(a)} Sliding-window occlusion sensitivity for Split Error Correction. The render shows the two segments being evaluated for merge over the depth-shaded mesh (top) or grayscale EM cutout (bottom); a $16\times16$\,px window is swept across the input at stride 4 (overlapping), zeroed, and the resulting drop in the classifier's probability $\lvert\Delta p\rvert$ is averaged onto each covered pixel and overlaid as a yellow$\rightarrow$red heat map (brighter red = larger drop = more important to the prediction), normalized per panel. \textbf{(b)} The same occlusion sensitivity for Merge Error Identification, where the input is a single, pre-split (merged) segment, shown in green. \textbf{(c)} Merge Error Correction: the split-mask-generation head applied to the same two exemplary merge-identification segments. Each pixel is colored by the predicted $P(\text{Neuron B})$, the dashed outline marks the ground-truth Neuron~B segment.
}
    \label{fig:interpretability}
\end{figure}

Finally, to understand and validate the model's learned heuristics, we investigated which image regions most strongly affected its answers. We found that the image regions with the greatest influence on the model’s judgments aligned broadly with our human assessment of their importance (Figure \ref{fig:interpretability}). For instance, for split error correction, we find that the model learns to pay attention to contact points between the two segments, while for merge error identification, it learns to pay attention to junction points. Moreover, when examining the model's proposed merge-error corrections, we find it both produces clear masks when confident about how to fix an error and flags regions that likely contain errors as unsure when it does not know how to split the segments apart.

\section{Calibration, Sharpness, \& Generalization: How well does the system know when it knows?}
\begin{figure}[h]
    \centering
    \includegraphics[width=\textwidth]{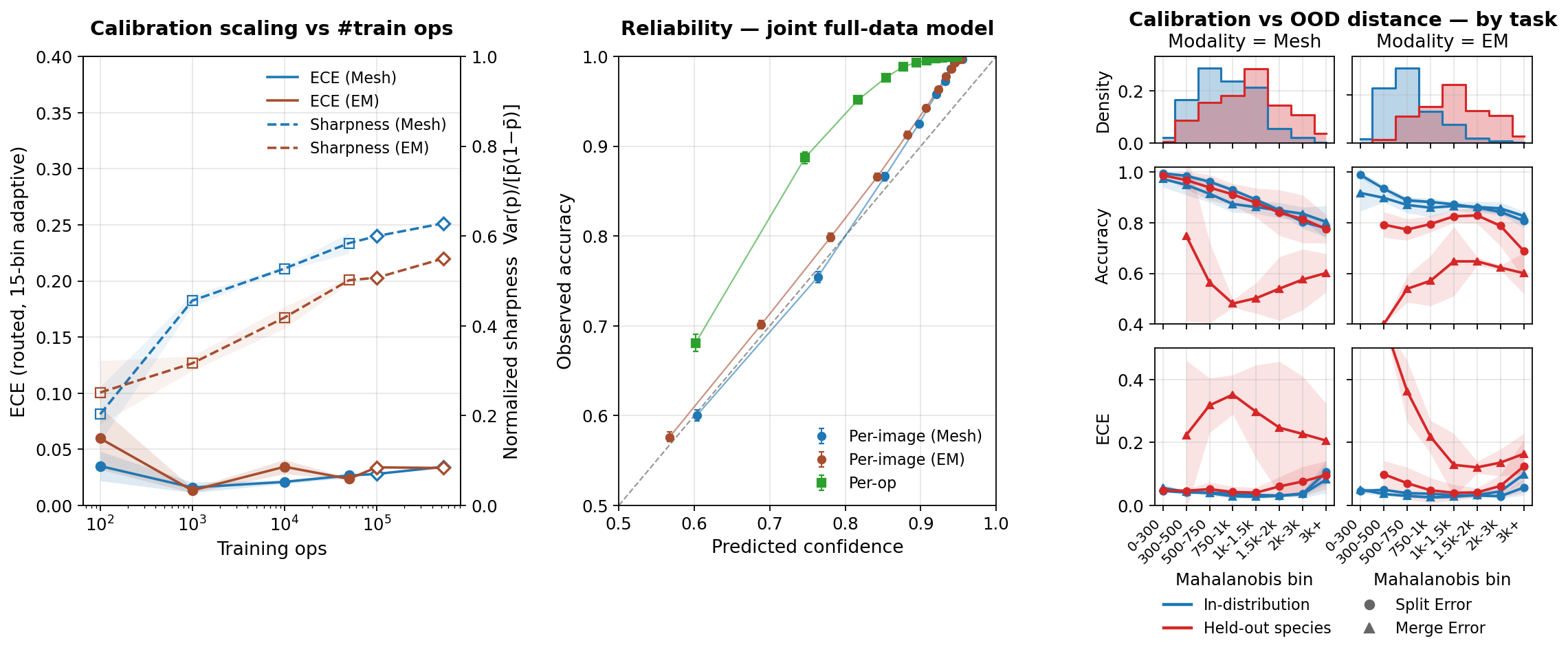}
    \caption{\textbf{(a)} Expected calibration error (ECE; 15-bin adaptive; left axis) and normalized sharpness $\mathrm{Var}(p)/[\bar p(1-\bar p)]$ (right axis) versus the number of training operations, by input modality (Mesh, EM). Bands are $\pm$1 SD across training seeds. Seeds per scale: 1 at all, 5 at 100, 3 at 1,000, 3 at 10,000, 2 at 50,000. Single-seed points are drawn as diamonds without bands. \textbf{(b)} Reliability curves for the joint classifier on the held-out test set. Each curve plots observed accuracy against predicted confidence ($\max(p, 1-p) \in [0.5, 1.0]$) in 12 equal-mass bins. Three traces: Per-image (Mesh), Per-image (EM), and Per-op (mean view probability per operation). Vertical bars are 95\% cluster-bootstrap confidence intervals on observed accuracy per bin. Diagonal = perfect calibration. \textbf{(c)} Calibration as a function of Mahalanobis distance to the in-distribution training-feature manifold, averaged across the four LOSO models. The 3$\times$2 grid shows: row 1 = density histogram of test-point Mahalanobis distances (heights normalized within (modality, split) for direct comparison); rows 2--3 = accuracy and ECE; columns = Mesh and EM. Four curves per acc/ECE sub-panel: in-distribution (blue) vs held-out species (red), each split by task drawn as the across-LOSO-model mean with a $\pm$1\,SD band. Merge-error curves exclude zebrafish/human (too few junction operations); the density histogram pools all species. 
}
    \label{fig:calibration}
\end{figure}

The second property we evaluate is calibration: how well a model's expressed confidence reflects its actual accuracy. Calibrated uncertainty is what allows a proofreading system to triage in deployment, applying confident predictions automatically and routing uncertain ones to human review.

We measure calibration using the adaptive expected calibration error (adaptive ECE) ~\cite{nixon2020measuringcalibrationdeeplearning}, which selects bin widths so that each bin contains the same number of samples. Calibration is strong for both mesh and imaging samples in the jointly trained model (Figure~\ref{fig:calibration}, Left). However, ECE alone can be misleading: a model that always predicts the base rate would achieve perfect calibration without making meaningful distinctions between samples. To capture this, we also report normalized sharpness $\mathrm{Var}(p) / [\bar{p}(1-\bar{p})]$, the variance of predictions normalized by the maximum achievable variance given the mean prediction~\cite{gneiting2007probabilistic}. We find that mesh and EM-based predictions show strong improvement in both calibration and sharpness with training scale.

At full data, calibration is strong overall: across both per-image modalities, the reliability curves track the diagonal closely, and the majority of samples sit in confidence bins where observed accuracy slightly exceeds predicted confidence. Notably, when we average predictions across the seven views available per sample, accuracy substantially exceeds the averaged confidence across every per-op bin. This indicates that the multi-view aggregation is underconfident relative to its true reliability. Modifying the model to fuse views internally rather than averaging post-hoc is a natural direction for improving per-sample calibration and sharpness.

Calibration as discussed so far is an in-distribution property. New connectomics datasets, however, will frequently be out-of-distribution relative to any training set, whether due to a new imaging modality, brain region, or species. We therefore evaluate how well a model can detect distributional shift and how its performance degrades as samples move further out of distribution. To do this, we run leave-one-species-out (LOSO) experiments: for each species, we train a model on the other three and evaluate on the held-out one.

We adapt the Mahalanobis distance OOD detection framework of Lee et al. to the LOSO setting, using class-conditional Gaussians fit on penultimate-layer features and compute the Mahalanobis distance (M-distance) for the held out species~\cite{mahalanobis2018generalized,lee2018simpleunifiedframeworkdetecting}. Across all four LOSO models, held-out species samples have a higher M-distance relative to in-distribution species.

As M-distance grows, accuracy decreases while calibration error increases. Breaking down the results by task, we find that when evaluating split-error correction using meshes, the in- and out-of-distribution curves track extremely closely across the full M-distance range, suggesting mesh-based geometric reasoning is highly generalizing for this task. For EM, by contrast, held-out-species performance is worse even at matched M-distance, indicating that the underlying imaging data does not transfer as cleanly. Merge-error identification fails to generalize under both modalities, but here the comparison is also the most demanding: because junction operations are too sparse in zebrafish and human, the merge curves effectively pit a mouse-trained model against fly (and vice versa), a pairing that spans an aggressive morphological shift.

\section{Efficiency: How quickly can the system adapt to new datasets?}
\begin{figure}[h]
    \centering
    \includegraphics[width=\textwidth]{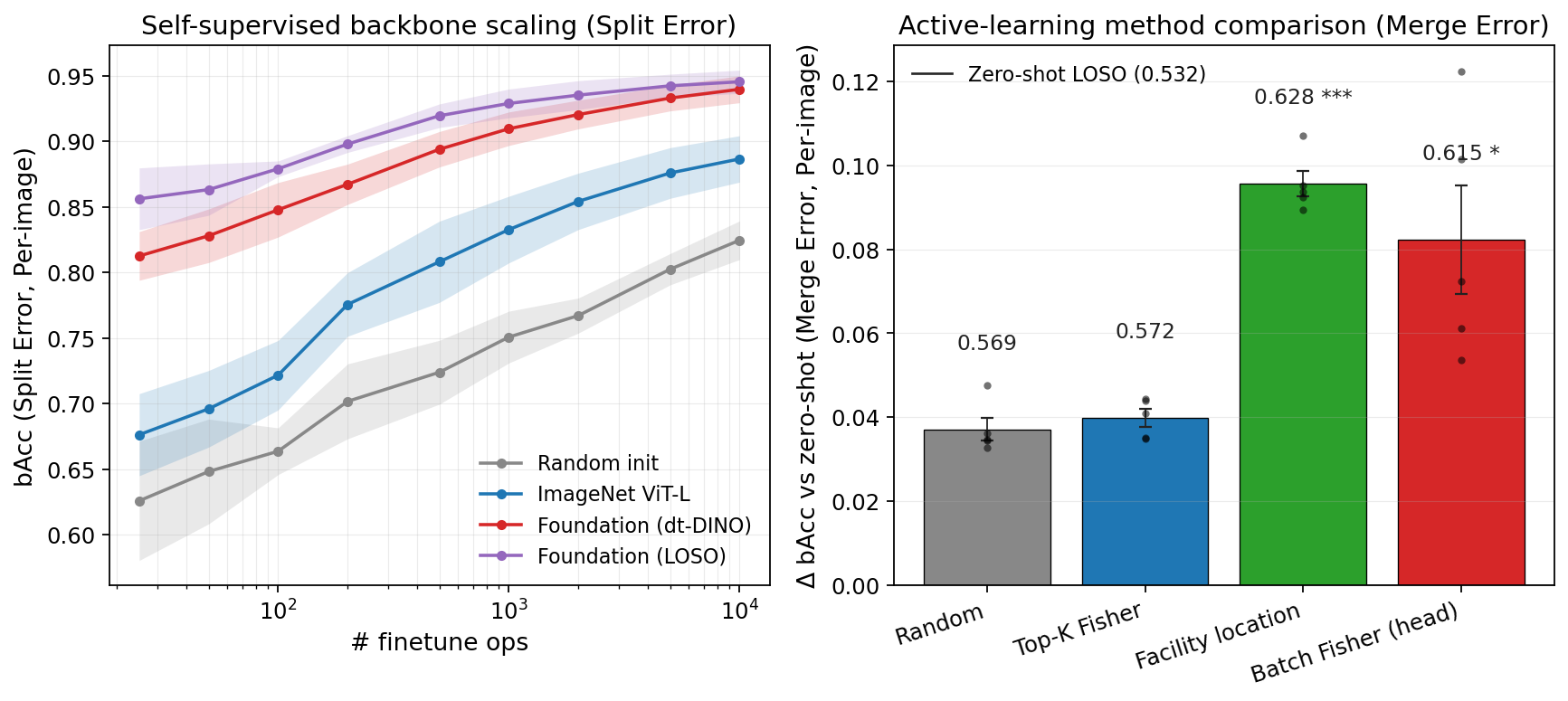}
    \caption{\textbf{(a)} Self-supervised backbone scaling on Split Error. Per-image balanced accuracy (bAcc) as a function of the number of finetune operations on a log $x$ axis. Each curve is the unweighted mean across the four species (mouse, fly, zebrafish, human); bands are $\pm$1 SE across species. Backbones: random init, ImageNet ViT-L, dt-DINO foundation model, and the LOSO foundation model (species held out at training time, evaluated on the held-out species).  \textbf{(b)} Active-learning method comparison on Merge Error at fixed budget $K=300$, LOSO-mouse held-out (the held-out species with the most zero-shot junction headroom; zero-shot bAcc $\approx$0.53). Bars show the change in held-out validation bAcc relative to zero-shot (annotated in the legend); absolute bAcc is shown above each bar. Mean $\pm$1 SE across 5 seeds (deterministic methods reuse identical selections; only optimizer / dropout RNG varies). Significance vs.\ Random by Welch's $t$-test: Top-K Fisher $p=0.444$ (n.s.); Facility location $p<0.001$ (***); Batch Fisher (head) $p=0.024$ (*).
}
    \label{fig:efficiency}
\end{figure}

The ability to extend to novel modalities and datasets is critical for deploying proofreading systems in the real world. Every new dataset will likely exhibit distributional shift relative to existing labeled data, and we have seen that strong performance requires substantial training data even within distribution. Some fine-tuning will therefore be necessary when extending to new datasets, and the question becomes how to do so efficiently: what conditions yield strong generalization with the fewest new annotated samples?

Two axes can be used to improve sample efficiency. The first is the quality of the model's representations: models with strong, transferable representations adapt more readily to new distributions \cite{yosinski2014transferable, kolesnikov2020bit}. The second is the choice of which samples to annotate. The active learning literature offers several criteria for ranking pool samples, including representativeness of the new distribution \cite{wei2015submodularity, sener2018active} and expected impact on the model via Fisher information \cite{ash2020deep, ash2021gone}. Our dataset supports evaluation along both axes.

To evaluate representations, we compare four pretraining strategies adapted to a held-out species: LOSO, self-supervised learning (SSL), ImageNet, and random initialization. LOSO uses the leave-one-species-out training procedure described earlier as a pretraining stage. For SSL, we evaluated a range of self-supervised objectives (see Appendix Figure ~\ref{fig:ssl-sweep}) and report the best-performing variant in Figure~\ref{fig:efficiency}; full results are in the Appendix. ImageNet and random initialization use the corresponding weights without modification. To isolate representation quality from downstream training capacity, all backbones are frozen and adaptation occurs only in two MLP heads attached for split error correction and merge error identification (mask generation is omitted from this analysis). We find that both pretraining strategies that leverage connectome-specific data (either unlabeled or labeled from another distribution) outperform ImageNet initialization. This indicates that pretraining on connectomics data yields representations that transfer to new species far more efficiently than pretraining on natural images.

To evaluate sample selection, we compare four active learning strategies operating on per-sample features: random sampling (baseline), top-k Fisher selection (rank by Fisher scalar with no diversity term), Fisher-weighted facility location (representativeness in feature space), and batch Fisher D-optimal on classifier-head gradients (parameter-space coverage in the head). Method definitions are detailed in the Appendix. We find that, at a fixed budget of $K=300$ ops on the LOSO-mouse held-out species, Facility location and Batch Fisher (head) both significantly outperform random selection on merge error identification. While absolute performance at this budget remains well below what would be needed to deploy a proofreading model on a new species, this benchmark provides the measurement infrastructure to compare active-learning strategies on connectomics proofreading at meaningful effect sizes.

\section{Conclusion}

Our benchmark has several limitations that future work could address. \textit{Rendering choices.} Our 2D slice-based imaging renders are lossy compared to full 3D EM cutouts, and better slice selection or learned 3D cutout embeddings could capture more information. \textit{Architectural coverage.} Our evaluations focus on vision transformers. 3D CNNs and point-cloud models are well-established in the proofreading literature and warrant comparison; the mesh and supervoxel data we collect would support a point-cloud formulation as a benchmark extension. \textit{Species imbalance.} Sample counts differ substantially across species, reflecting differences in proofreading abundance and per-neuron data size. Cross-species generalization claims should be read with this imbalance in mind, particularly for LOSO experiments on smaller held-out species.

Within these limits, we have shown that modern ML practices transfer to connectomic proofreading. A unified image-based formulation lets a single ViT handle split correction, merge identification, and merge-correction mask generation across four species at human-level accuracy; performance scales with data; joint training matches or exceeds species-specific models; in-distribution calibration is strong; Mahalanobis distance provides a label-free signal of distributional shift; connectomics pretraining substantially outperforms ImageNet for adapting to new species; and active learning further reduces adaptation cost. The benchmark provides the measurement infrastructure for the field to extend these results.

\begin{ack}
ESB acknowledges HHMI, Lisa Yang, NIH R01AG087374, NIH 1R01EB024261, NIH 1R01AG070831. JB acknowledges funding support from the Fannie and John Hertz Foundation, and compute support from Modal Labs through the Academic Grant Program. TF acknowledges funding support from the German Academic Scholarship Foundation and the McGovern Institute. 
\end{ack}

\clearpage
\printbibliography

\clearpage
\appendix

\begin{figure}
\centering
\includegraphics[width=1\linewidth]{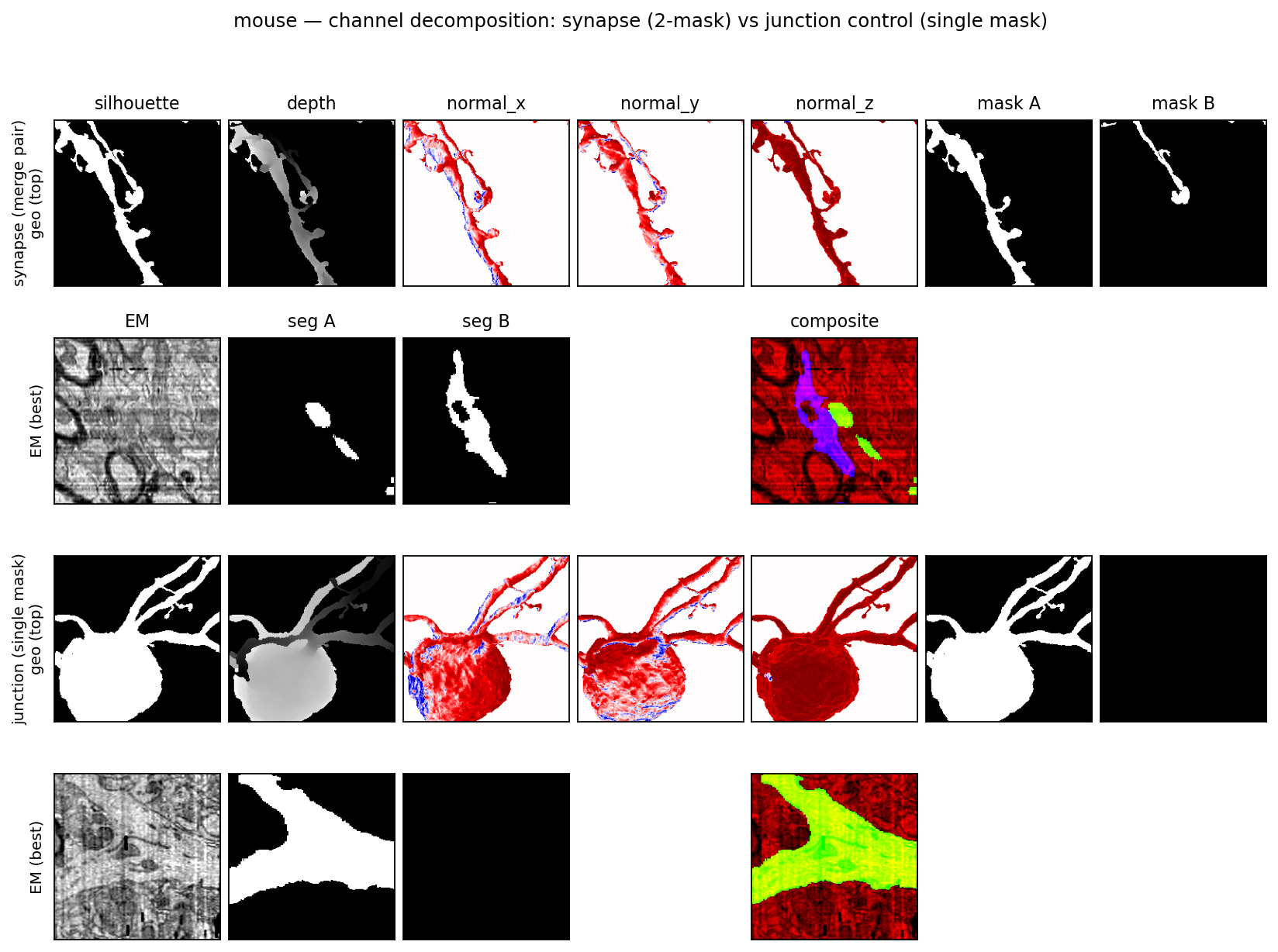}
\caption{Channel Decomposition of Mouse Training Samples. Two examples are shown: a synapse pair with two populated segment masks and a junction control in single-mask form, where only mask A is populated. Each sample includes a 7-channel top-view geometry render and the corresponding best-plane imaging (EM) slice, decomposed into raw EM intensity, segment masks A/B, and an RGB composite.}
\label{fig:geo-em-data-example-rendering}
\end{figure}

\section{Data Generation Details}
\subsection{Connectomics Data Sources}
\label{sec:appendix-data-sources}
\begin{figure}
    \centering
    \includegraphics[width=0.75\linewidth]{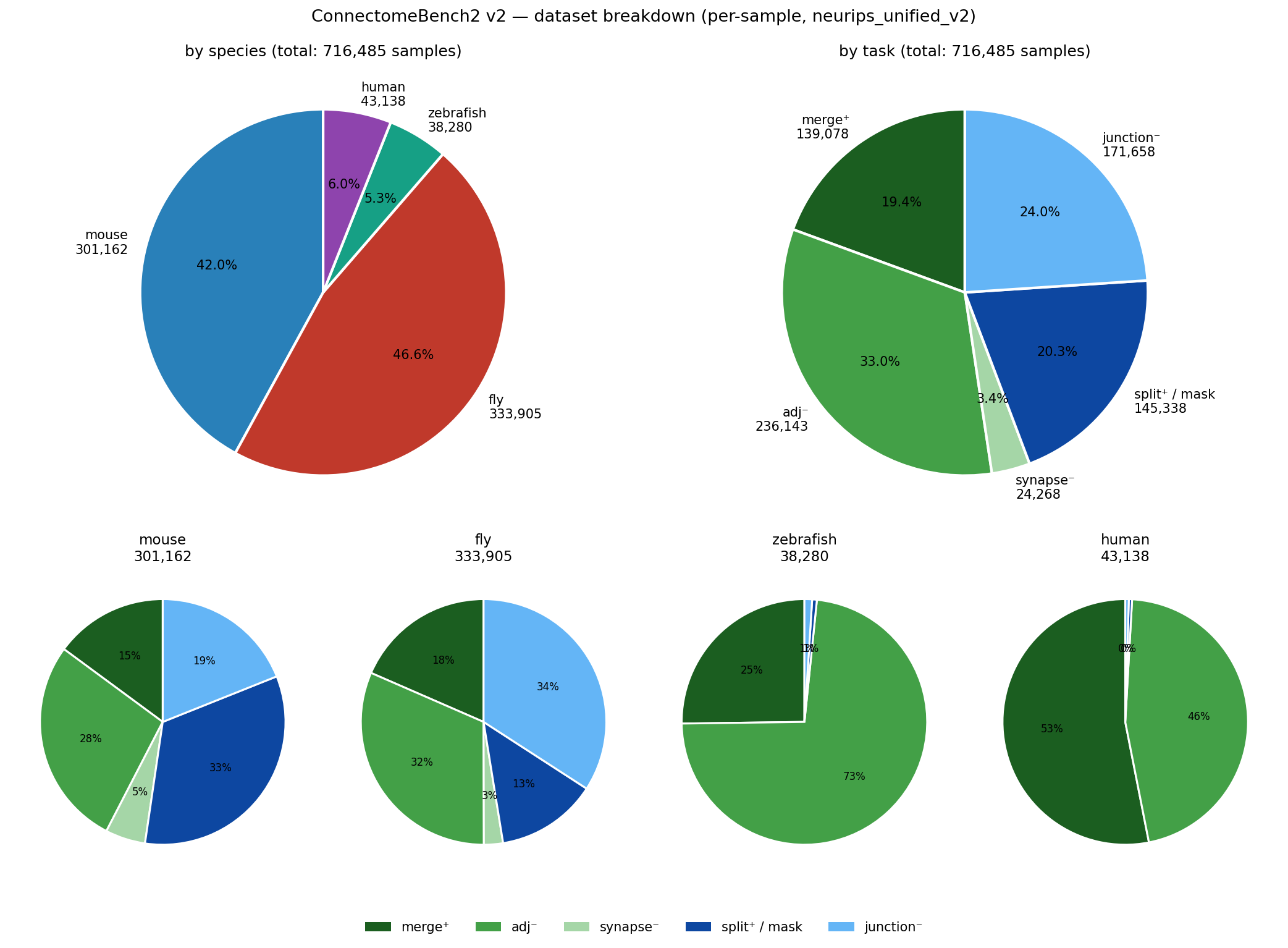}
    \caption{Data distribution by species and task. While within task, labels are roughly balanced, species are imbalanced due to different availabilities of edits and controls.}
    \label{fig:dataset_sizes}
\end{figure}


\begin{table}[ht]
  \centering
  \small
  \begin{tabular}{@{}lllrlp{3cm}@{}}
  \toprule
  \textbf{Project} & \textbf{Species} & \textbf{Volume} &
   \textbf{Neurons} & \textbf{Proofread fraction} &
  \textbf{Reference} \\
  \midrule
  FlyWire & \textit{D. melanogaster} (adult) & whole
  brain &
  $\sim$139{,}255 & $\sim$100\% & Dorkenwald et al.,
  \textit{Nature} 2024 \\
  \addlinespace
  MICrONS & mouse (visual cortex) & $\sim$1\,mm$^3$
  &
  $>$200{,}000  & $\sim$1\% & MICrONS
  Consortium, \textit{Nature} 2025 \\
  \addlinespace
  Fish1 & larval zebrafish & whole larval
  brain &
  $\sim$100,000 & <1\% & Svara et al., \textit{Nat. Methods}
   2022 \\
  \addlinespace
  H01 & human (temporal cortex) & $\sim$1\,mm$^3$
  fragment & $\sim$50{,}000 & <1\% & Shapson-Coe et al.,
  \textit{Science} 2024 \\
  \bottomrule
  \end{tabular}
  \caption{Major EM connectome datasets used in this
  work.}
  \label{tab:connectomes}
  \end{table}

Roots are nodes in an append-only, directed proofreading graph. Each proofreading edit adds new root nodes and directed provenance edges: merge corrections connect two or more roots to a single new root, while split corrections connect one previous root to two or more new roots.

At a given timestamp, the final leaf roots ('descendant roots') of this graph (i.e. roots with no outgoing provenance edges at that time) represent the latest segmentation/proofreading state. 
  
\subsection{Synapse controls}
\label{sec:appendix-synapses}

 \begin{table}[h]
  \centering
  \small
  \begin{tabular}{lllll}
  \toprule
  Species & Project & Synapse table & Filter & Reference
  \\
  \midrule
  Mouse & MICrONS & \texttt{synapses\_pni\_2} &
  \texttt{size} $> 2000$ & \cite{turner2020synaptic} \\
  Fly   & FlyWire & \texttt{synapses\_nt\_v1} &
  \texttt{cleft\_score} $\geq 50$ &
  \cite{buhmann2021automatic,heinrich2018synaptic} \\
  \bottomrule
  \end{tabular}
  \caption{CAVE synapse tables used as merge-candidate
  sources. 
  Pairs additionally required pre-synapse root id $\neq$ post-synapse root id to exclude autapses (i.e. rare synapses of neurons onto themselves), with the pre-synapse root drawn from the species' proofread-neuron table.}
\end{table}

\subsection{Junction controls}
\label{sec:appendix-junctions}

These are used as negative labels for the false merge identification task. 

While edits, synapse controls and adjacent segment controls are reliable by construction (human curated), junctions are inherently less trustworthy as it is harder to identify the absence of a false merge than the correctness of two segments being separate. We thus only used roots from the proofread root tables in mouse/fly and from the most proofread set of roots of human, zebrafish  (due to their less advanced proofreading state), and manually inspected a subset of them for their validity.

For each sampled root we download its skeleton (simplified graph representations of their rough morphology and topology), we then use skeleton nodes with degree >2 as junction sites. 

\paragraph{Zebrafish and human root selection} As zebrafish and human were proofread significantly less comprehensively at the time of writing, junction controls pose a problem as they're not directly inferable from edits alone (only absence of edits in a comprehensively proofread neuron establishes presence of a true junction as opposed to a merge error). Thus, only junctions from roots above a certain size ( $\geq$100um for zebrafish,  $\geq$500um for human) with proofreading densities comparable to fly/mouse (>20 ops/mm for zebrafish,  >10 ops/mm for human) were included, yielding just a few hundred controls samples for these two species, reflecting their early proofreading state.

\subsection{Data Rendering}
\label{sec:appendix-data-rendering}
Because the same proofreading site can be represented either as the pre-correction segmentation or the post-correction segmentation, we rendered both single-segment and two-segment variants where applicable, preventing the model from using trivial mesh/rendering artifacts associated with graph state rather than local morphology.

To make the model more invariant to spatial scale, and to avoid leakage from edit point placement, extents and exact center points were randomly jittered once at render time. Cutout sizes and geometry rendering extent ranges were chosen manually per-species to balance local detail with global context.

\begin{table}[t]
\centering
\caption{Per-species rendering jitter and cutout scales.}
\label{tab:rendering_scales}
\begin{tabular}{@{}lccc@{}}
\toprule
\textbf{Species} & \textbf{Center jitter} & \textbf{Geometry extent} & \textbf{Imaging cutout} \\
                 & \textbf{(nm)}          & \textbf{(nm)}             & \textbf{(nm)} \\
\midrule
Human     & 300 & 6{,}000--12{,}000 & 2{,}500 \\
Mouse     & 300 & 5{,}000--10{,}000 & 2{,}500 \\
Fly       & 300 & 2{,}000--4{,}000  & 1{,}500 \\
Zebrafish & 300 & 2{,}500--5{,}000  & 1{,}500 \\
\bottomrule
\end{tabular}
\end{table}

\section{Model Training}
\subsection{Tasks and Sample Sources}

\paragraph{Augmentations.} 
\label{sec:appendix-training-augmentation}
During training we apply the following
  augmentations; validation and test use clean inputs.

  \textit{Spatial} (applied jointly to input and label mask):
  \begin{itemize}
      \item Horizontal flip, $p = 0.5$
      \item Rotation, angle $\sim \mathcal{U}(-25^\circ, 25^\circ)$
      \item Translation, $(\Delta x, \Delta y) \sim \mathcal{U}(-0.15, 0.15)
  \cdot (W, H)$
  \end{itemize}

  \textit{Photometric} (input-only, channel-aware: applied to depth + normals for
   geometry inputs, and to the grey intensity channel for EM inputs; silhouette
  and mask channels are left untouched):
  \begin{itemize}
      \item Brightness jitter, factor $\sim \mathcal{U}(0.6, 1.4)$
      \item Contrast jitter, factor $\sim \mathcal{U}(0.6, 1.4)$
      \item Gaussian blur, $p = 0.4$, $\sigma \sim \mathcal{U}(0.1, 2.0)$
  \end{itemize}

  \textit{Occlusion / dropout} (input-only):
  \begin{itemize}
      \item Random erasing, $p = 0.4$, box size $\sim \mathcal{U}(0.1, 0.3) \cdot
   (H, W)$, zeroed across all channels
      \item Channel-group dropout (geometry only), $p = 0.15$ independently per
  group: silhouette, depth, and normals (the three normal components are dropped
  together to preserve unit-norm semantics)
  \end{itemize}

\subsubsection{Task Routing Based on Sample Type}
\label{sec:appendix-task-routing}

\begin{table}[h!]
\centering
\small
\caption{Learning tasks and corresponding positive and negative samples.}
\label{tab:tasks_samples}
\begin{tabularx}{\linewidth}{@{}>{\raggedright\arraybackslash}X
                                >{\raggedright\arraybackslash}X
                                >{\raggedright\arraybackslash}X@{}}
\toprule
\textbf{Task} & \textbf{Positive samples} & \textbf{Negative samples} \\
\midrule

Binary classification of merge corrections (of false splits) (dual segment)
&
\textit{Proofreader merge corrections}: two segments (with split error) that were merged into one segment
&
\textit{Synapse controls}: neurons (final proofread segments) that synapse onto each other 

\textit{Adjacent controls}: segments spatially adjacent to edits
\\
\midrule

Binary classification of false merges (single segment)
&
\textit{Proofreader split corrections}: one segment (with merge error) that was split into two segments 
&
\textit{Junction controls}: skeleton junctions on proofread neurons \\
\midrule

Mask segmentation for split correction (of false merges) (single segment)
&
\textit{Proofreader split corrections}: one segment (with split error) that was split into two segments (post-split segments used as labels) 
&
--- \\

\bottomrule
\end{tabularx}
\end{table}

\subsection{Compute}
\label{sec:appendix-reproducibility}

\paragraph{Architecture details.} The 7-channel geometry patch embedding is
a $\mathrm{Conv2d}(7, D, k{=}16, s{=}16)$. From an ImageNet-pretrained
3-channel backbone, RGB filters are copied into the normal channels
(\texttt{nx, ny, nz}); silhouette, depth, mask$_A$, mask$_B$ are
Kaiming-initialized. The 3-channel EM patch embedding is a separate
$\mathrm{Conv2d}(3, D, 16, 16)$ initialized directly from the ImageNet RGB
filters. Each classification head is $\text{Dropout}(0.1) +
\text{Linear}(D, 2)$. The mask decoder reads the $14\!\times\!14$ patch
tokens and upsamples to $224\!\times\!224$ via four
$\mathrm{ConvTranspose2d}(s{=}2)$ stages
($D \!\to\! 256 \!\to\! 128 \!\to\! 64 \!\to\! 2$, BatchNorm + ReLU between
stages). A learned 3-way view embedding (front/side/top) is added to the
patch grid before the decoder.

\paragraph{Loss masking and gating.} For each sample, the per-sample
endpoint, junction, and synapse cross-entropies are masked by indicators
that the sample's task matches the head and (for synapse) that the row
carries a synapse label, then averaged over matching samples in the batch.
$\mathcal{L}_{\text{mask}}$ is gated to rows with valid dual-mesh
ground-truth (split-correction edits with a separately rendered post-edit
geometry); EM rows and merge-side rows contribute zero gradient to this
term. CE and Dice in the permutation-invariant mask loss are computed on
foreground pixels only; samples with no foreground contribute zero. For
non-split rows (single-segment foreground) the loss reduces to the
single-channel form.

\paragraph{Optimizer and schedule.} AdamW
($\beta_1{=}0.9, \beta_2{=}0.999$, weight decay $0.05$ for ViT-B and $0.1$
for ViT-L). Reference learning rate is $1\!\times\!10^{-4}$ at batch
size 32 with square-root scaling, giving $2.83\!\times\!10^{-4}$ at the
single-GPU batch size 256 and $4.90\!\times\!10^{-4}$ at the DDP batch
size 768. The backbone receives an additional $0.1{\times}$ LR multiplier
relative to the heads, mask decoder, and EM patch embedding. The schedule
is 5 epochs of linear warmup (start factor $0.1$) followed by cosine
annealing to zero. Total epochs are 20 for the peak run and 25--40 for
low-data scaling and species ablations. Drop-path is $0.1$ (ViT-B) /
$0.25$ (ViT-L), head dropout $0.1$, label smoothing $0.1$, gradient norm
clipped at $1.0$, mixed precision in bfloat16 (avoids attention overflow
on ViT-L).

\paragraph{Class balance and species blending.} A
\texttt{WeightedRandomSampler} is constructed per epoch from per-cell
(species$\times$task) sample weights computed under a \texttt{sqrt}
blending strategy: each cell is weighted by $\sqrt{N_{\text{cell}}}$ then
spread uniformly within the cell, multiplied by the per-archive weight (mouse and
human edits $1.0$, fly $0.7$, zebrafish $0.5$; junction and synapse
archives $1.0$). Sample weights are additionally inverted by per-task
class frequency. Junction (false-merge) controls are hardcoded to
\emph{same\_neuron=True}; synapse controls to \emph{same\_neuron=False}.

\paragraph{Augmentation.} Spatial augmentations are applied jointly to the
input and any mask label: horizontal flip ($p{=}0.5$), in-plane rotation
$\pm 25^\circ$ (nearest-neighbor for masks), and translation
$\pm 15\%$ of image size. Input-only photometric augmentations are
channel-aware (so unit-norm normal vectors are not corrupted): brightness
$[0.6, 1.4]$ and contrast $[0.6, 1.4]$ on the depth + normals channels for
geometry and on the EM grey channel for imaging; Gaussian blur
($p{=}0.4$, $\sigma \in [0.1, 2.0]$) on the same intensity channels;
random erasing ($p{=}0.4$, box size $10$--$30\%$) over all input channels;
and channel-group dropout ($p{=}0.15$ each for silhouette, depth, and the
normals triplet) on geometry views.

\paragraph{Hardware.}
\begin{itemize}
    \item \emph{Training:} H100 80\,GB containers,
    1--3 GPUs per job (the largest configuration is ViT-L on
    3$\times$H100 via DDP). 48--64 host CPU cores, 1.5\,TB ephemeral disk
    for archive extraction. Per-job wall-clock cap: 24\,h.
    \item \emph{Inference:} A100 40\,GB containers, single
    GPU, batch size 256, bfloat16 autocast. The full test-split inference
    sweep over $\sim$30 checkpoints in
    Table~\ref{tab:peak_performance} completes in under 4 GPU-hours per
    checkpoint.
\end{itemize}

\paragraph{Seeds.} Default seed is 42 (controls model init, sampler,
permutation, and \texttt{torch.cuda} RNG). Scaling-curve points use
seeds $\{0,1,2,3,4\}$: 5 seeds at $N{=}100$ ops, 3 seeds at $N{=}1\text{k}$
and $N{=}10\text{k}$, 2 seeds at $N{=}50\text{k}$, 1 seed at the full data
point. Seeds for individual runs are recorded in \texttt{registry.yaml}.

\paragraph{Hyperparameter selection.} ViT defaults (LR, weight decay,
warmup, label smoothing, drop-path, gradient clip) follow the
\texttt{ResNetTrainingConfig.apply\_model\_defaults} bridge, kept fixed
across all reported runs in the main text. Loss-weight ratios
($\lambda_{\text{ep}}{:}\lambda_{\text{jn}}{:}\lambda_{\text{mask}}{:}\lambda_{\text{syn}} = 1{:}1{:}1{:}0.25$),
the mask CE/Dice mixing weight ($w_{\text{CE}}{=}0.5$), and per-archive
blend weights were chosen from preliminary experiments on a smaller
3-species blend and \emph{not retuned} for the final 4-species runs
reported here. Active-learning hyperparameters and finetune budgets are
described in Appendix~[ref to AL appendix]; for each AL method we hold all
other training settings fixed at the values above.

\subsection{Mask Predictions to Supervoxel Splits}\label{sec:appendix-mask-to-split}

The mask head produces a 2-channel logit map per view ($224\!\times\!224$,
arbitrary A/B labeling). Converting this to an applicable proofreading
edit requires a partition of the local supervoxel graph, which we obtain
in three stages: 3D hull reconstruction from the per-view masks,
seed-point sampling from the hulls, and multicut on the supervoxel graph.

\paragraph{Stage 1 -- Hull reconstruction.} We support two
reconstruction paths:
\begin{itemize}
    \item \emph{Visual hull} (\texttt{build\_visual\_hull}): The three
    orthographic masks (front/side/top) are independently unprojected
    along their respective camera axes onto a $32$--$256^3$ voxel grid
    (resolution adapts to \texttt{inner\_size\_nm}). The 3D occupancy is
    the AND of the three view-axis sweeps. Each segment ($A$ and $B$) is
    reconstructed independently. This path uses no segmentation data.
    \item \emph{Seg-constrained hull}
    (\texttt{build\_seg\_constrained\_hull}): Each view's $A$ and $B$
    mask casts \emph{rays} (3D planes swept along the camera axis) into a
    real CAVE segmentation cutout fetched at \texttt{seg\_mip=0} and
    cached locally. A voxel votes into hull $A$ if it is (i) inside the
    foreground supervoxel volume \emph{and} (ii) hit by an $A$-mask ray
    from a sufficient number of views (default $\geq 3$, with optional
    pixel dilation). Hull $B$ is built analogously. Voxels in both hulls
    are removed.
\end{itemize}
The seg-constrained variant is the production path; the visual-hull
variant is a baseline that does not require segmentation access.

\paragraph{Auto-alignment of A/B across views.} Mask-channel labels are
arbitrary per view (the permutation-invariant mask loss never enforces a
global $A/B$ identity). Before voting, we therefore try all four flip
combinations of side and top relative to front (the front view fixes the
reference) and select the assignment that maximizes
$\min(|H_A|, |H_B|)$, the size of the smaller hull. This favors balanced
A/B partitions and is what the \texttt{flipped} field in the per-sample
JSON records.

\paragraph{Stage 2 -- Seed sampling.} We sample $k$ random points (default
$k{=}5$, optionally drawn from a range \texttt{k\_range} on retry)
uniformly from the occupied voxels of hull $A$ and hull $B$, converting
voxel centers back to nm-space coordinates. These become the source and
sink seeds for the multicut.

\paragraph{Stage 3 -- Supervoxel multicut.} Source and sink seeds are
passed to \texttt{compute\_split} (the wrapper around pychunkedgraph's
\texttt{run\_multicut}) along with a window of
$1.5\!\times\!\texttt{inner\_size\_nm}$. The result is a tuple
$(\hat{C}_{\text{src}}, \hat{C}_{\text{snk}}, \hat{E}_{\text{cut}})$
of source/sink supervoxel components and the cut edges between them.
Failed runs (\texttt{Invalid vertex index} errors, multicut timeouts) are
retried up to \texttt{n\_retries} times with fresh random seed-point
draws from the same hulls.

\paragraph{Ground-truth comparison.} For every sample we also run
\texttt{compute\_split} on the human-placed seeds stored in the parquet
metadata (\texttt{meta["sources"]}, \texttt{meta["sinks"]}), producing
$(C^{\text{GT}}_{\text{src}}, C^{\text{GT}}_{\text{snk}},
E^{\text{GT}}_{\text{cut}})$. Both the GT and predicted multicuts run
inside the same subprocess on the same fetched chunked-graph, so they
share supervoxel IDs.

\paragraph{Metric: supervoxel IoU.} Because the predicted/GT
\emph{source} and \emph{sink} labels are themselves arbitrary, we score
the partition label-invariantly:
\[
\mathrm{sv\_iou}
\;=\;
\max\!\Big(
  \tfrac{1}{2}\big[
    \mathrm{IoU}(\hat{C}_{\text{src}}, C^{\text{GT}}_{\text{src}})
    + \mathrm{IoU}(\hat{C}_{\text{snk}}, C^{\text{GT}}_{\text{snk}})
  \big],\;
  \tfrac{1}{2}\big[
    \mathrm{IoU}(\hat{C}_{\text{src}}, C^{\text{GT}}_{\text{snk}})
    + \mathrm{IoU}(\hat{C}_{\text{snk}}, C^{\text{GT}}_{\text{src}})
  \big]
\Big),
\]
the maximum over the two label assignments of the average source/sink
component IoU. We treat $\mathrm{sv\_iou} > 0.8$ as the binary
\emph{good-split} threshold.

\begin{figure}[h]
    \centering
    \includegraphics[width=0.7\textwidth]{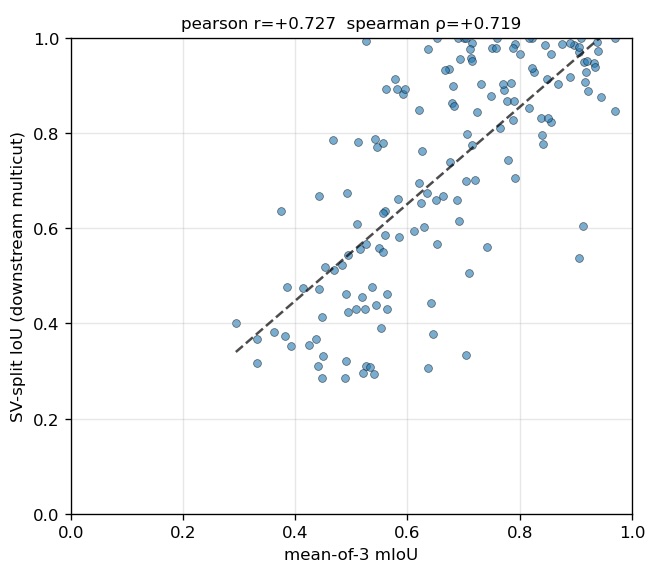}
    \caption{\textbf{Mask IoU predicts supervoxel-split quality.}
    Per-sample $\mathrm{sv\_iou}$ (3D supervoxel IoU between the mask-driven
    multicut and the human-seed multicut, label-invariant) versus
    $\mathrm{mask\_iou\_mean}$ (per-view label-invariant 2D mask IoU,
    averaged across front/side/top). Light blue: $n{=}94$ split-OK
    junction samples. Black: bucketed mean $\pm 1$ standard deviation.
    Pearson $r{=}0.671$, Spearman $\rho{=}0.422$. The bucketed mean
    rises monotonically from $0.43$ at $\mathrm{mask\_iou}\!\approx\!0.25$
    to $0.95$ at $\mathrm{mask\_iou}\!\approx\!0.98$.}
    \label{fig:sv-iou-vs-mask-iou}
\end{figure}

\paragraph{2D mask IoU as a leading indicator.} For a sanity check that
upstream mask quality predicts downstream split quality, we additionally
compute the per-view label-invariant 2D IoU
(\texttt{compute\_pred\_gt\_mask\_iou}) between the predicted and GT
masks --- for each view, taking the max over both A$\leftrightarrow$A /
A$\leftrightarrow$B label assignments and reporting the per-sample mean
across the three views. This is the \texttt{mask\_iou\_mean} statistic
plotted in Figure~\ref{fig:sv-iou-vs-mask-iou}.

\paragraph{Results.}
Figure~\ref{fig:sv-iou-vs-mask-iou} plots $\mathrm{sv\_iou}$ against
$\mathrm{mask\_iou\_mean}$ across $n{=}94$ split-OK junction samples.
The two are positively correlated (Pearson $r{=}0.727$, Spearman
$\rho{=}0.719$), and the bucketed mean rises monotonically. Two practical implications follow. First, the per-view 2D mIoU
reported in the main accuracy table is a meaningful proxy for proofreading
utility, not just a pixel-level fit metric: high-mIoU samples almost
always produce supervoxel partitions that recover most of the
ground-truth cut. Second, the residual scatter at high mask IoU (some
samples at $\mathrm{mask\_iou}{>}0.9$ still landing at
$\mathrm{sv\_iou}{<}0.5$) shows that pixel-level mask quality is
necessary but not sufficient: a mask that gets the boundary roughly right
in 2D can still seed multicut on the wrong side of a thin process if the
hull sample lands in an ambiguous voxel.

\clearpage
\newpage

\subsection{Foundation Model: Architecture, Pretraining, and SSL Sweep}\label{sec:appendix-foundation-model}

\paragraph{Architecture details.} For pre-training, we use a dual-tower architecture, where each tower is an independent ViT dedicated to one input modality (geometry, or imaging data).

Each tower represents a multi-view architecture, and follows the same three-stage layout: a per-view ViT-L encoder, an attention-pooled aggregator across views, and a stack of self-supervised heads attached both to the aggregated scene-level representation, and each view's representation (CLS tokens).

\paragraph{Inputs.} To generate samples for self-supervised training, we sample 'anchor points' spaced along proofread roots. For each we render 9 mesh views ($3$ elevation angles $\times$ $3$ zoom extents at $224\!\times\!224$ pixels) and stack them into a 7-channel tensor (silhouette, depth, three normal components $n_x, n_y, n_z$, and two per-segment binary masks). An aligned EM cutout at the same anchor feeds the parallel tower as a 3-channel tensor (intensity + two per-segment masks).

\paragraph{Pretraining task.} For every anchor, three scenes are
generated by modifying the input geometry while holding the underlying
neuron identity fixed:
\begin{itemize}
    \item \textsc{Clean}. The mesh remains intact. To match the channel
    layout of the corrupted classes, the visible mesh is partitioned into halves $A$ and $B$ by a deterministic PCA-aligned plane through the anchor; the two halves populate the per-segment mask channels.
    \item \textsc{Synthetic merge}. A real adjacent root is discovered in a small segmentation cutout around the anchor. Both root meshes are fetched, cropped to a window centered on the interface point, and rendered jointly. The mask channels separately label segment-$A$ and segment-$B$ pixels. Pairs are filtered by species-specific size thresholds to avoid merging sub-segment dust.
    \item \textsc{Synthetic split}. The local mesh is cut by a uniformly random plane through the anchor. Faces on one side are discarded, the cut surface is capped with fan-triangulated faces so the result appears closed, and only the largest connected component is retained.
    Branches intersecting the plane fall away as small disconnected
    components, producing a realistic ``missing branch'' appearance.
\end{itemize}
The EM tower receives the same 3-class label on its corresponding EM
crops, with the per-segment masks supplying the same partitioning
information in image space.

\paragraph{Objective.} Per tower, three losses are summed jointly from
step 0:

\begin{center}
\begin{tabular}{lp{0.6\linewidth}}
\toprule
Term & What it supervises \\
\midrule
\texttt{corruption\_cls} ($\times 2$ towers) & Scene-level: was the geometry corrupted, and how. \\
\texttt{DINO}            ($\times 2$ towers) & View-consistency at the scene CLS, two random views per scene. \\
\texttt{iBOT}            ($\times 2$ towers) & Local visual structure at masked patch tokens, mask ratio $0.30$. \\
\bottomrule
\end{tabular}
\end{center}

The corruption-classification term provides a scene-level semantic
signal; DINO and iBOT supply the canonical self-distillation pair (cross-view at the CLS, masked-patch within-view at the patch tokens). Both self-distillation losses use a frozen EMA teacher with cosine momentum annealed from $0.996$ to $1.0$, prototype centering, student temperature $0.1$, and teacher temperature $0.04$. The prototype layer is gradient-zeroed for the first epoch, following the original DINO recipe. Loss weights are $1.0 / 1.0 / 0.5$ for corruption classification, DINO, and iBOT respectively, identical across both towers; the iBOT down-weighting follows the original iBOT recipe. Both towers contribute, so the total objective consists of six loss terms.

\paragraph{Augmentation.} A small augmentation stack is applied to mesh
views during training only. The most notable component is a
\emph{geometry-coherent erosion}: a max-pool over the inverse silhouette
shrinks the visible support, and all geometry channels (silhouette,
depth, normals) are coherently masked by the eroded support. This
simulates blobby or under-segmented mesh boundaries and prevents
overfitting to the razor-sharp silhouettes produced by the offscreen
renderer. Erosion is applied uniformly across all three corruption
classes with identical probability so it cannot leak label information.

\paragraph{Training.} Optimization uses AdamW with base learning rate
$1\!\times\!10^{-4}$ linearly scaled by the global effective batch size
(reference batch 256), weight decay $0.05$, and a 60-epoch schedule with
5-epoch linear warmup followed by cosine annealing. We use gradient
accumulation $4\times$, max gradient norm $1.0$, bfloat16 autocast, and
DDP across $8\!\times\!\text{H100}$ GPUs; wall-clock time for the
canonical ViT-L run less than 24 hours. Both towers train
jointly in a single optimization loop; the student backbone forward is
executed once per tower per step, and the resulting per-view CLS is
reused by both the pooler+classification head and the DINO head, avoiding
a doubled activation footprint. The pretraining corpus is 70k scenes
across five species (mouse, fly, human, zebrafish, LICONN), root-
stratified 90/10 into train/val.

\paragraph{Downstream protocol.} For all evaluations using
foundation-model features in this paper, the mesh tower (backbone +
attention pooler) is frozen and used as a feature extractor. A small
probe head --- LayerNorm $\to$ Linear $\to$ GELU $\to$ Dropout $\to$
Linear --- is trained per task on the pooled scene embedding $z$. The
label-efficiency curves in Section 6, including the
$N_{\text{train}} \times \text{seed}$ sweeps across mouse, fly,
zebrafish, and human on both endpoint-correction and junction-error
tasks, all share this protocol; only the upstream backbone differs
(random init, ImageNet-supervised, our dual-encoder DINO/iBOT
pretraining, and a leave-one-species-out supervised reference).


\subsubsection*{Self-Supervised Ablation Sweep}\label{sec:appendix-ssl-sweep}

\begin{figure}[h]
    \centering
    \includegraphics[width=\textwidth]{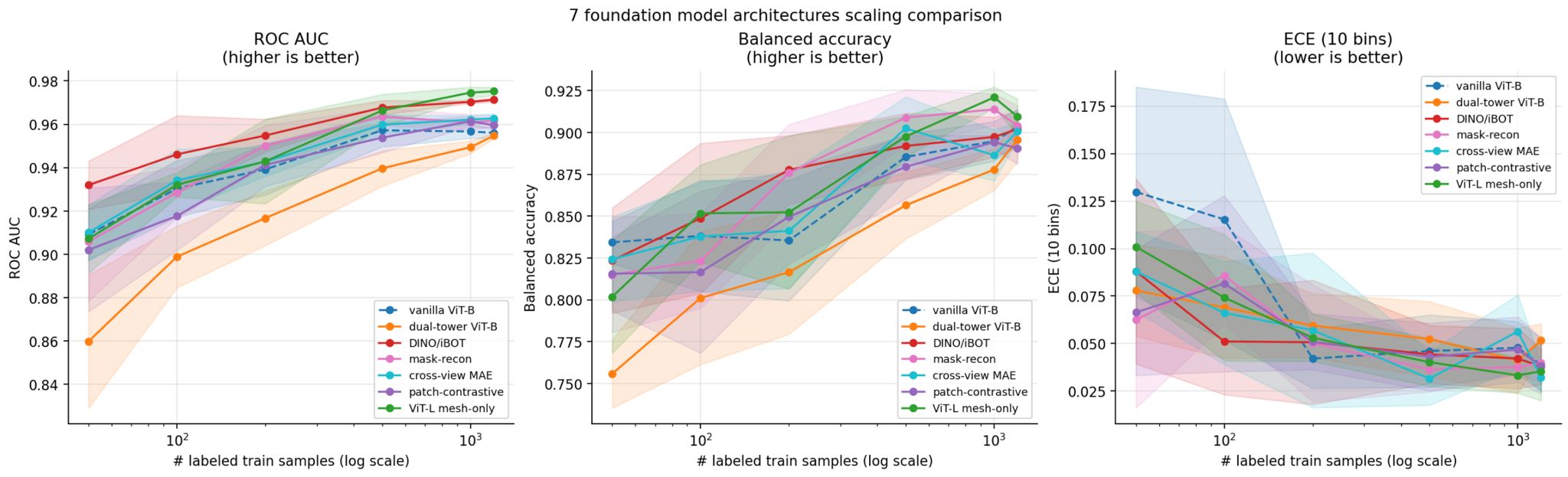}
    \caption{\textbf{Self-supervised pretraining sweep.} Seven
    architectures pre-trained on the same 70k 5-species mesh corpus for
    30 epochs, frozen, and evaluated with a small MLP probe on the mouse
    endpoint-correction + junction-identification task. Panels: ROC-AUC
    (higher better), balanced accuracy (higher better), 10-bin ECE
    (lower better) as a function of labelled training samples (log
    scale). Curves are means over 5 random splits; bands are
    $\pm 1$ SD. \emph{ViT-L mesh-only} leads in the high-data regime;
    \emph{DINO/iBOT} leads in the low-label regime; the remaining ViT-B
    variants trail both across all three metrics. The headline
    foundation model in the main text combines the ViT-L backbone, the
    DINO/iBOT objective, and the dual-encoder architecture.}
    \label{fig:ssl-sweep}
\end{figure}

Figure~\ref{fig:ssl-sweep} compares seven self-supervised architectures,
each pre-trained on the same 70k 5-species mesh corpus for 30 epochs,
then frozen and evaluated using a small MLP probe on the mouse
endpoint-correction $+$ junction-identification task. Five of the
architectures share a ViT-B/16 encoder and differ only in pretraining
recipe; the remaining two vary encoder size (ViT-L mesh-only) or encoder
number (dual-tower ViT-B). The three panels report ROC-AUC, balanced
accuracy, and 10-bin ECE as functions of the number of labelled training
samples (log scale). Each curve is the mean over 5 random splits with
shaded $\pm 1$ standard-deviation bands.

\paragraph{Implementation of the seven recipes.} Five share a common
scaffolding --- a ViT-B/16 encoder, an attention-based view pooler that
combines the nine orthographic mesh views into a scene embedding, a
3-class corruption-classification head trained against pairwise
clean-vs-corrupted labels, and a 256-dim projection head trained with
multi-view InfoNCE ($\tau=0.07$, in-batch negatives) --- and differ only
in what they add on top.
\begin{itemize}
    \item \emph{vanilla ViT-B}: scaffolding only.
    \item \emph{DINO/iBOT}: replaces InfoNCE with self-distillation
    against an EMA teacher copy of the entire network (cosine momentum
    $0.996 \!\to\! 1.0$); a 3-layer DINO head projects each CLS token
    onto 4096 prototypes ($\tau_s{=}0.1, \tau_t{=}0.04$); a parallel iBOT
    head is applied per-patch on a $30\%$-masked view.
    \item \emph{mask-recon}: adds a per-patch linear decoder that
    reconstructs the two binary mask channels under BCE-with-logits on
    merge/split samples whose masks were zeroed by the existing
    mask-dropout regularizer.
    \item \emph{cross-view MAE}: adds a 6-layer transformer decoder
    (dim 384) that, given one view masked at $75\%$, cross-attends to a
    clean second view of the same scene and reconstructs the geometry
    channels under per-patch z-normalised L2.
    \item \emph{patch-contrastive}: adds a small per-patch projection
    head and a dense InfoNCE objective between geometrically corresponding
    patches in two same-axis views at different zoom levels;
    correspondences are computed exactly from the orthographic camera
    geometry, and patches below $5\%$ silhouette fraction are filtered
    out.
    \item \emph{ViT-L mesh-only}: vanilla recipe with the encoder swapped
    for ViT-L/16 (1024-dim).
    \item \emph{dual-tower ViT-B}: departs from the shared scaffolding
    --- two independently weighted ViT-B encoders, one on 7-channel mesh
    views and one on 3-channel EM views, each with its own pooler,
    classifier, and projection head, trained jointly under four
    equally-weighted losses (mesh corruption, mesh InfoNCE, EM corruption,
    EM InfoNCE) with no shared parameters or cross-modal terms.
\end{itemize}
All seven models are pretrained for 30 epochs with AdamW, a 5-epoch
warmup followed by cosine decay, and gradient clipping at $1.0$;
auxiliary objectives that supplement the shared base (iBOT, mask-recon,
cross-view MAE, patch-contrastive) are weighted at $0.5$ relative to the
two base losses.

\paragraph{Findings.} Once labels become abundant (beyond
$\sim\!10^3$ samples), all seven architectures converge toward a narrow
performance band, with balanced accuracy in roughly $0.875$--$0.92$.
\emph{ViT-L mesh-only} (green) achieves the strongest performance in the
high-data regime, while \emph{DINO/iBOT} (red) performs best in the
low-label regime; the remaining ViT-B variants consistently trail these
two across all three metrics. The headline foundation model used in the
main text combines these two findings --- a ViT-L backbone with the
DINO/iBOT objective --- and adopts the dual-encoder design from
\emph{dual-tower ViT-B} to accommodate the EM modality, although the
dual-tower variant on its own underperforms the single-tower ViT-B at
this scale (consistent with the doubled parameter count outpacing the
30-epoch budget in the sweep).

\subsection{Active-Learning Formalism}\label{sec:appendix-active-learning}

This section formalizes the four sample-selection methods compared in
Figure~\ref{fig:efficiency}b and the head-only finetune protocol used to
evaluate them. All methods operate on a pool of \emph{per-op} embeddings,
where the views of an operation are first averaged so a unit of selection
corresponds to one annotation request.

\paragraph{Pool construction.} Let $\mathcal{D}_{\text{pool}}$ be the
held-out species' validation cubes, sliced to the species being evaluated
in Figure~\ref{fig:efficiency}b (LOSO-mouse). For each operation $i$ we
forward each of its $V_i$ views through the frozen LOSO encoder
(\texttt{model.\_encode}), capture the CLS token $\phi_{i,v}$ and the
routed-task logits $\ell_{i,v} \in \mathbb{R}^2$ (the endpoint head for
split-correction operations, the junction head for false-merge
operations), and average across views:
\[
x_i = \tfrac{1}{V_i}\sum_{v=1}^{V_i} \phi_{i,v}, \qquad
\ell_i = \tfrac{1}{V_i}\sum_{v=1}^{V_i} \ell_{i,v}.
\]
Probabilities and the per-sample binary Fisher scalar are then
\[
p_i = \mathrm{softmax}(\ell_i) \in \Delta^1, \qquad
f_i \;=\; 2\, p_{i,0} \, p_{i,1} \in [0, \tfrac{1}{2}].
\]
$f_i$ is the variance of a Bernoulli with parameter $p_{i,1}$ (up to a
factor of 2) and equivalently the Fisher information of the binary head
output with respect to its scalar logit; samples on the decision boundary
($p_{i,1} = 0.5$) maximize $f_i$. The
\emph{Fisher-weighted} variant we use in facility location additionally
incorporates the squared feature norm,
$w_i = f_i \cdot \|x_i\|_2^2$, so that uncertain samples in
high-magnitude regions of feature space dominate the coverage objective.

\paragraph{Selection budgets and per-task split.} We evaluate budgets
$K \in \{10, 30, 100, 300\}$ in the artifacts; the main-text bar plot
reports $K{=}300$. The pool contains both endpoint and junction
operations, and the LOSO-mouse pool is roughly balanced across the two
tasks. We split $K$ approximately 50/50 between tasks and reallocate
slack if one pool can't fill its share (\texttt{run\_loso\_selection.py:\_split\_k\_by\_task});
each method is then run independently on each task slice and the two
selections are concatenated.

\paragraph{Method 1 -- Random.} Uniform sampling without replacement
from $\mathcal{D}_{\text{pool}}$ at each seed:
$\mathcal{S}_{\text{rand}} \sim \mathrm{Uniform}\binom{[N]}{K}$. This is
the only method whose selection itself depends on the seed; the other
three are deterministic given the pool (see Seeds and variance below).

\paragraph{Method 2 -- Top-$K$ Fisher.} Rank by the Fisher scalar and
keep the top $K$:
\[
\mathcal{S}_{\text{topK}} \;=\; \mathrm{argsort}_i^{(K)}\, f_i.
\]
This is the simplest uncertainty-only baseline, with no diversity term;
we include it as a control to isolate the effect of representativeness
from raw Fisher magnitude.

\paragraph{Method 3 -- Fisher-weighted facility location.} Greedy
maximization of a Fisher-weighted facility-location objective on cosine
similarities of penultimate features. Let
$\hat{x}_i = x_i / \|x_i\|_2$ and
$s_{ij} = \hat{x}_i^\top \hat{x}_j$. The facility-location value of a
selected set $\mathcal{S}$ is
\[
V(\mathcal{S}) \;=\; \sum_{i=1}^{N} w_i \cdot \max_{j \in \mathcal{S}}\, s_{ij},
\]
with the maximum interpreted as $-\infty$ when $\mathcal{S}$ is empty.
$V$ is monotone submodular in $\mathcal{S}$, so greedy yields a
$(1-1/e)$ approximation to the optimum. We grow $\mathcal{S}$ one sample
at a time, picking at each step
\[
j^\star = \arg\max_{j \notin \mathcal{S}} \;\sum_i w_i \cdot \max\!\big(\, s_{ij} - c_i, 0 \,\big),
\qquad c_i = \max_{j' \in \mathcal{S}}\, s_{ij'},
\]
i.e.\ the candidate with the largest \emph{marginal} weighted coverage
gain over what is already covered (\texttt{select\_facility\_location} in
\texttt{\_al\_selection.py}).

\paragraph{Finetune protocol.} For each (method, $K$, seed) we restore
the LOSO-mouse checkpoint, freeze the backbone and the
synapse/mask heads, and train only the routed cls heads
(\texttt{cls\_endpoint}, \texttt{cls\_junction}) with AdamW
($\mathrm{lr} = 10^{-5}$, weight decay $0.05$), label smoothing $0.1$,
and a fixed budget of $1000$ gradient steps regardless of $K$ ---
small-$K$ runs cycle the loader and see proportionally more passes, so
every (method, $K$) gets the same compute budget. Loss is masked
cross-entropy as in §Methods (synapse and mask terms are not active
under the AL routing). The finetuned checkpoint is then evaluated on the
LOSO-mouse held-out validation cubes via per-op mean-prob aggregation,
matching the main-text evaluation protocol; we report
\emph{junction-task balanced accuracy} on the held-out validation set as
the headline metric in Figure~\ref{fig:efficiency}b (LOSO-mouse has the
largest zero-shot junction headroom, $\approx 0.532$, of the four
LOSO conditions).

\paragraph{Seeds and variance.} We run 5 seeds per (method, $K$) cell.
Random is the only method whose \emph{selection} varies across seeds;
the other three are deterministic given the pool, so for those methods
the seed varies only the optimizer / dropout RNG of the finetune
loop, not which $K$ ops are labeled. Significance tests in the main
text caption (Welch's $t$-test vs.\ Random at $K{=}300$) treat each
seed as one observation and use unequal variances.

\section{Supplemental Results}

\subsection{Results by species, task, and confidence using ViT-B}
\begin{table}[ht]
\centering
\caption{Per-species peak performance (vitb\_full, the headline joint model): per-sample Split / Merge Error balanced accuracy and mask mIoU (per-image mean and per-op best). 95\% cluster-bootstrap CIs on op\_id.}
\label{tab:peak_per_species}
\begin{tabular}{lcccc}
\toprule
Species & Split Error bAcc (\%) [95\% CI] & Merge Error bAcc (\%) [95\% CI] & mIoU [95\% CI]\\
\midrule
Mouse & 98.6 [98.4, 98.7] & 95.7 [95.4, 96.0] & 0.716 [0.713, 0.719] \\
Fly & 96.5 [96.3, 96.8] & 85.5 [85.0, 86.0] & 0.638 [0.633, 0.643] \\
Zebrafish & 94.0 [93.2, 94.8] & 87.6 [77.8, 95.5] & 0.633 [0.546, 0.722] \\
Human & 95.9 [95.3, 96.4] & 95.0 [87.5, 100.0] & 0.718 [0.640, 0.782]  \\
\bottomrule
\end{tabular}
\end{table}

\begin{table}[ht]
\centering
\caption{Per-species peak (vitb\_full): per-sample sensitivity (TPR) and specificity (TNR) for Split / Merge Error at threshold 0.5 (bAcc $=\tfrac12$(TPR+TNR)). 95\% cluster-bootstrap CIs on op\_id.}
\label{tab:peak_per_species_tpr_tnr}
\begin{tabular}{lccc}
\toprule
Species & Task & TPR (\%) [95\% CI] & TNR (\%) [95\% CI] \\
\midrule
Mouse & Split Error & 98.3 [98.0, 98.6] & 98.8 [98.6, 99.0] \\
Mouse & Merge Error & 97.0 [96.8, 97.3] & 94.4 [93.9, 94.9] \\
Fly & Split Error & 95.9 [95.5, 96.3] & 97.2 [96.9, 97.4] \\
Fly & Merge Error & 73.9 [72.9, 74.9] & 97.2 [96.9, 97.4] \\
Zebrafish & Split Error & 91.1 [89.4, 92.5] & 96.9 [96.4, 97.5] \\
Zebrafish & Merge Error & 89.5 [73.3, 100.0] & 85.7 [73.0, 96.9] \\
Human & Split Error & 95.3 [94.5, 96.1] & 96.4 [95.7, 97.1] \\
Human & Merge Error & 90.0 [75.0, 100.0] & 100.0 [100.0, 100.0] \\
\bottomrule
\end{tabular}
\end{table}

\begin{table}[ht]
\centering
\caption{Per-species peak (vitb\_full), HIGH-CONFIDENCE only: per-sample TPR/TNR restricted to predictions with confidence $\max(p,1-p)\geq0.9$ (the selective-prediction view, as in Fig.~3 calibration). Coverage = fraction of ops retained at that threshold. 95\% cluster-bootstrap CIs on op\_id}
\label{tab:peak_per_species_tpr_tnr_conf90}
\begin{tabular}{lcccc}
\toprule
Species & Task & Coverage (\%) & TPR (\%) [95\% CI] & TNR (\%) [95\% CI] \\
\midrule
Mouse & Split Error & 49.8 & 99.6 [98.6, 100.0] & 100.0 [100.0, 100.0] \\
Mouse & Merge Error & 43.4 & 99.9 [99.9, 100.0] & 99.9 [99.7, 100.0] \\
Fly & Split Error & 39.7 & 99.7 [99.3, 99.9] & 99.9 [99.9, 100.0] \\
Fly & Merge Error & 34.7 & 99.0 [98.4, 99.7] & 99.9 [99.8, 100.0] \\
Zebrafish & Split Error & 45.5 & 90.5 [84.3, 96.4] & 100.0 [100.0, 100.0] \\
Zebrafish & Merge Error & 5.6 & 100.0 [100.0, 100.0] & 100.0 [100.0, 100.0] \\
Human & Split Error & 34.9 & 99.9 [99.6, 100.0] & 99.9 [99.7, 100.0] \\
Human & Merge Error & 23.3 & 100.0 [100.0, 100.0] & 100.0 [100.0, 100.0] \\
\bottomrule
\end{tabular}
\end{table}
\FloatBarrier
\newpage
\subsection{Merge Error Correction Examples}
\begin{sidewaysfigure}
    \centering
    \includegraphics[width=0.8\textwidth]{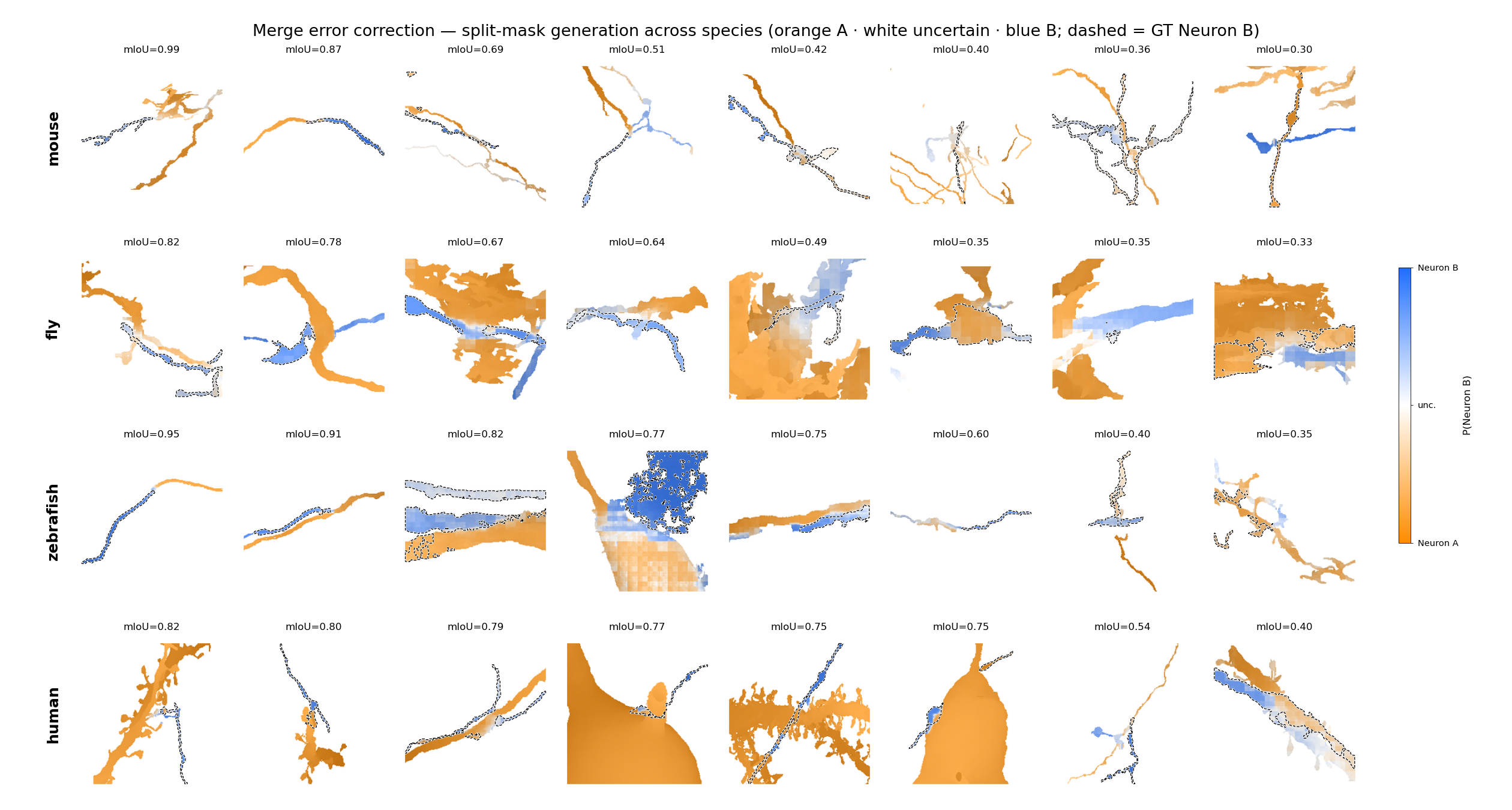}
    \caption{Merge error correction examples.}
    \label{fig:mask-examples}
\end{sidewaysfigure}

%

\clearpage
\newpage

\end{document}